\documentclass[journal]{IEEEtran}
\pdfoutput=1

\usepackage[utf8]{inputenc}

\usepackage{cite}

\usepackage[pdftex]{graphicx}
\graphicspath{{./images/}}
\DeclareGraphicsExtensions{.pdf}

\usepackage[cmex10]{amsmath}
\usepackage{amsfonts}
\usepackage{bbm}
\usepackage{mathtools}
\usepackage{algorithm}
\usepackage{algpseudocode}
\usepackage{paralist}
\usepackage{import}

\usepackage{amsthm}
\usepackage{url}
\usepackage[hidelinks]{hyperref} 
\usepackage{aliascnt}
\newtheorem{theorem}{Theorem}
\newtheorem{lemma}{Lemma}

\newtheorem{remark}{Remark}

\usepackage[flushleft]{threeparttable}

\usepackage{bibentry}

\usepackage{comment}

\usepackage{xspace}

\ifCLASSOPTIONcompsoc
  \usepackage[caption=false,font=normalsize,labelfont=sf,textfont=sf]{subfig}
\else
  \usepackage[caption=false,font=footnotesize]{subfig}
\fi

\usepackage{adjustbox}

\usepackage{booktabs}
\usepackage{ifthen}

\usepackage{xparse}

\ifCLASSOPTIONdraftcls
    \usepackage[mathlines]{lineno}
    \usepackage{xcolor}

    \linenumbers
    \newcommand\getcurrentref[1]{%
     \ifnumequal{\value{#1}}{0}%
      {??}%
      {\the\value{#1}}%
    }
    \newcommand{\parnum}{\P\arabic{parcount}}
    \newcounter{parcount}
    \usepackage{etoolbox}
    \newcommand{\paragraphnumbering}{%
        \everypar={%
            \stepcounter{parcount}%
            \setcounter{linenumber}{1}%
            \leavevmode\marginpar[\parnum\hfill]{\parnum}}%
    }
    \renewcommand{\section}[1]{%
        \refstepcounter{section}%
        \everypar={}\nolinenumbers%
        {\normalfont\normalsize\centering\scshape \thesection.~#1\\}
        \marginpar[\S\getcurrentref{section}\hfill]{\S\getcurrentref{section}}%
        \setcounter{parcount}{0}%
        \linenumbers\paragraphnumbering%
    }
    \renewcommand{\subsection}[1]{%
        \everypar={}\nolinenumbers%
        \refstepcounter{subsection}%
        {\noindent\normalfont\normalsize\itshape \thesubsectiondis~{#1}}%
        \linenumbers\paragraphnumbering}
    \renewcommand{\subsubsection}[1]{%
        \refstepcounter{subsubsection}%
        \thesubsubsectiondis~{\itshape #1}:%
    }
\else%
\fi

\newif\ifShowNotes

\ShowNotestrue

\ifShowNotes
    \usepackage{color}
    \newcommand{\note}[1]{{\color{red}{\bf Note:}\ #1}}
    
\else
    \newcommand{\note}[1]{ }
\fi

\usepackage{cleveref}
\crefname{section}{Section}{Sections}
\crefname{algorithm}{Algorithm}{Algorithms}
\crefname{equation}{Equation}{Equations}
\crefname{figure}{Figure}{Figures}
\crefname{table}{Table}{Tables}
\crefname{theorem}{Theorem}{Theorems}
\crefname{lemma}{Lemma}{Lemmas}
\newcommand{\crefrangeconjunction}{--}

\usepackage[normalem]{ulem}

\newif\ifShowStandardDeviation

\newcommand{\rvaluesd}[2]{\ifthenelse{\equal{#2}{}}{#1}{#1 $\pm$ #2}}

\ifShowStandardDeviation
    \newcommand{\rvalue}[2]{\ifthenelse{\equal{#2}{}}{#1}{#1 $\pm$ #2}}
\else
    \newcommand{\rvalue}[2]{#1}
\fi

\title{Network Unfolding Map by Vertex-Edge Dynamics Modeling}

\author{
    Filipe~Alves~Neto~Verri, 
    Paulo~Roberto~Urio, 
    and~Liang~Zhao,~\IEEEmembership{Senior~member,~IEEE}%
    \thanks{
      © 2018 IEEE. Personal use of this material is permitted. Permission from IEEE must
      be obtained for all other uses, in any current or future media, including
      reprinting/republishing this material for advertising or promotional purposes,
      creating new collective works, for resale or redistribution to servers or lists, or
      reuse of any copyrighted component of this work in other works.
      Published version: \texttt{http://ieeexplore.ieee.org/document/7762202/}.
      DOI 10.1109/TNNLS.2016.2626341.
    }%
    \thanks{
        This research was supported by the São Paulo State Research Foundation
        (FAPESP), the Coordination for the Improvement of Higher Education
        Personnel (CAPES), and the Brazilian National Research Council (CNPq).
    }%
    \thanks{
        F.A.N.~Verri and P.R.~Urio are with the Institute of Mathematical and Computer
        Sciences (ICMC), University of São Paulo, São Carlos, SP, Brazil.  Email:
        \{filipeneto,urio\}@usp.br.
    }%
    \thanks{
      L.~Zhao is with Faculty of Philosophy, Sciences, and Letters at Ribeirão Preto
      (FFCLRP), University of São paulo, Ribeirão Preto, SP, Brazil. Email: zhao@usp.br.
    }
}

\ifCLASSOPTIONpeerreview
\else
\fi

\newcommand{\dd}{\mathop{}\!\mathrm{d}}

\renewcommand{\vec}[1]{\mathbf{#1}}

\newcommand{\vertexnotation}[1]{$v_{#1}$\xspace}
\newcommand{\vertex}[1]{\vertexnotation{#1}}
\newcommand{\vertices}[2]{\vertexnotation{#1} and~\vertexnotation{#2}}
\newcommand{\betweenvertices}[2]{\vertices{#1}{#2}}

\newcommand{\edgenotation}[2]{\mbox{$(#1, #2)$}\xspace}
\newcommand{\edge}[2]{edge~\edgenotation{#1}{#2}}
\newcommand{\edges}[4]{edges~\edgenotation{#1}{#2} and~\edgenotation{#3}{#4}}

\newcommand{\timenotation}[1]{\mbox{$#1$}\xspace}
\renewcommand{\time}[1]{time~\timenotation{#1}}

\newcommand{\class}[1]{class~\mbox{$#1$}\xspace}

\newcommand{\systemnotation}[1]{\mbox{$#1$}\xspace}
\newcommand{\system}[1]{system~\systemnotation{#1}}

\newcommand{\systems}[2]{systems~\systemnotation{#1} and~\systemnotation{#2}}

\newcommand{\particle}[1]{particle~$p$\xspace}

\newcommand{\suchthat}{|}
\newcommand{\card}[1]{\left\lvert#1\right\rvert}
\newcommand{\LabeledSet}{\mathcal{L}}
\newcommand{\UnlabeledSet}{\mathcal{U}}
\newcommand{\VertexSet}{\mathcal{V}}
\newcommand{\EdgeSet}{\mathcal{E}}

\newcommand{\ModelName}{Labeled Component Unfolding system\xspace}
\newcommand{\ModelNameFirst}{Labeled Component Unfolding (LCU) system\xspace}
\newcommand{\ShortModelName}{LCU system\xspace}
\newcommand{\SimulationModelName}{Labeled Component Unfolding\xspace}
\newcommand{\WDynamicsName}{walking\xspace}
\NewDocumentCommand\MaxSet{m+g}{%
    \IfNoValueTF{#2}%
        {\max\!\left\{#1\right\}}%
        {\max\!\left\{#1,\;#2\right\}}%
}
\NewDocumentCommand\MinSet{m+g}{%
    \IfNoValueTF{#2}%
        {\min\!\left\{#1\right\}}%
        {\min\!\left\{#1,\;#2\right\}}%
}
\newcommand{\prob}[1]{\Pr\!\left[#1\right]}

\newcommand{\cprob}[2]{\Pr\!\left[#1 \middle\suchthat #2\right]}
\newcommand{\expect}[1]{\mathrm{E}\!\left[#1\right]}
\newcommand{\cexpect}[2]{\mathrm{E}\!\left[#1 \middle\suchthat #2\right]}
\newcommand{\DominationName}{cumulative domination\xspace}
\newcommand{\oDomination}[2][ij]{\tilde{\delta}^{#2}_{#1}}
\newcommand{\oDMatrix}[1][c]{\tilde{\Delta}^{#1}}
\newcommand{\Domination}[2][ij]{\delta^{#2}_{#1}}
\newcommand{\DMatrix}[1][c]{\Delta^{#1}}
\newcommand{\submission}[1][c]{\sigma^{#1}_{ij}}
\newcommand{\RelativeCurrentSubmission}[1][c]{\tilde{\sigma}^{#1}_{ij}}
\newcommand{\CurrentDomination}[2][ij]{\tilde{n}^{#2}_{#1}}

\newcommand{\FinalTrans}{P^c}
\newcommand{\finaltrans}[1][ij]{p^c_{#1}}

\newcommand{\SurvivalParam}{\lambda}

\newcommand{\Sourceness}{\rho^c_i}
\newcommand{\SourceSet}[1][c]{\mathcal{G}^c}

\newcommand{\newparticles}[1][i]{g^c_{#1}}
\newcommand{\onewparticles}[1][i]{\tilde{g}^c_{#1}}
\newcommand{\vnewparticles}[1][c]{\vec{g}^{#1}}

\newcommand{\onpart}[1][ij]{\tilde{n}^c_{#1}}
\newcommand{\ovpart}[1][c]{\vec{\tilde{n}}^c}
\newcommand{\npart}[1][ij]{n^c_{#1}}
\newcommand{\xnpart}[2][ij]{n^{#2}_{#1}}

\newcommand{\vpart}[1][c]{\vec{n}^{#1}}
\newcommand{\Npart}[1][c]{N^{#1}}

\DeclareMathOperator*{\argmax}{arg\,max}

\newcommand{\diag}{\operatorname{diag}}
\newcommand{\Degree}{\operatorname{deg}}

\newcommand{\defeq}{\coloneqq}

\newcommand{\ParticleMoved}[2][t+1]{\mathrm{I}_{ij}(#2,~#1)}
\newcommand{\AParticleMoved}[1][ij]{\mathrm{I}^c_{#1}}
\newcommand{\ofinaltrans}[2][ij]{\prob{\AParticleMoved[#1](#2) = 1}}
\newcommand{\cofinaltrans}[1]{\cprob{\AParticleMoved[ij](t+1)=1}{\RelativeCurrentSubmission(t)=#1}}

\newcommand{\RealSet}{\mathbb{R}}
\newcommand{\BigO}{\mathcal{O}}
\newcommand{\Complexity}[1]{\BigO\!\left(#1\right)}

\newcommand{\kNN}{{\itshape k}-NN\xspace}

\begin{document}

\ifCLASSOPTIONpeerreview
    \IEEEpeerreviewmaketitle
\else
    \maketitle
\fi


\begin{abstract}
The emergence of collective dynamics in neural networks is a mechanism of the
animal and human brain for information processing.  In this paper, we develop a
computational technique using distributed processing elements in a complex network, which are called
particles,
to solve semi-supervised learning problems. Three actions
govern the particles' dynamics: generation, walking, and absorption.
Labeled vertices generate new particles that compete against rival particles for
edge domination. Active particles randomly walk in the network until they are
absorbed by either a rival vertex or an edge currently dominated by rival
particles.
The result from the model evolution consists of sets of edges arranged by the
label dominance.  Each set tends to form a connected subnetwork to represent a
data class. Although the intrinsic dynamics of the model is a stochastic one, we
prove there exists a deterministic version with largely reduced
computational complexity; specifically, with linear growth. Furthermore,
the edge domination process corresponds to an unfolding map in such way that edges
``stretch'' and ``shrink'' according to the vertex-edge dynamics. Consequently,
the unfolding effect
summarizes the relevant relationships between vertices and the uncovered data
classes.  The proposed model captures important details of connectivity patterns
over the vertex-edge dynamics evolution, in contrast to previous approaches
which
focused on only vertex or only edge dynamics.
Computer simulations reveal that the new model can identify nonlinear features in
both real and artificial data, including boundaries between distinct classes and
overlapping structures of data.
\end{abstract}



\newcommand{\termdef}[1]{\mbox{\emph{#1}}}

\section{Introduction}


\IEEEPARstart{S}{emi-supervised} learning (SSL) is one of the machine learning
paradigms, which lies
between the unsupervised and supervised learning paradigms.  In SSL problems,
both unlabeled
and labeled data are taken into account in class or cluster formation and
prediction processes~\cite{Chapelle2006,Zhu2009}. In real-world applications, we
usually have partial knowledge on a given dataset. For example, we certainly do
not know every movie actor except a few famous ones; in a large-scale
social network, we just know some friends; in biological domain, we are far away
from completely obtaining a figure of the functions of all genes, but we know the
functions of some of them. Sometimes, although we have a complete or almost
complete knowledge of a dataset, labeling it by hand is lengthy and expensive.
So it is necessary to restrict the labeling scope. For these reasons, partially
labeled datasets are often encountered. In this sense, supervised and
unsupervised learning can be considered as extreme and special cases of
semi-supervised learning. Many semi-supervised learning techniques
have been developed, including generative models~\cite{NigamEtal2000},
discriminative models~\cite{Loog2015},
clustering and labeling techniques~\cite{WagstaffEtal2001},
multi-training~\cite{ZhouLi2005}, low-density separation
models~\cite{Vapnik1998}, and graph-based
methods~\cite{Thiago2012c,Cheng2014,Zhang2015}.
Among the approaches listed above, graph-based SSL has triggered much
attention. In this case, each data instance is represented by a vertex and is
linked to other vertices according to a predefined affinity rule. The labels are
propagated to the whole graph using a particular optimization
heuristic~\cite{BelkinEtal2006}.

\termdef{Complex networks} are large-scale graphs with nontrivial topology~\cite{NewmanBarabasiWatts2006}.
Such networks introduce a powerful tool to describe the interplay of topology, structure, and dynamics of
complex systems~\cite{NewmanBarabasiWatts2006,Newman2010}. Therefore, they provide a groundbreaking mechanism
to help us understand the behavior of many real systems.
Networks also turn out to be an important mechanism for data representation and analysis~\cite{SilvaZhao2016}.
Interpreting data sets as \mbox{complex networks} grant us to access the
inter-relational nature of data items further.
For this reason, we consider the network-based approach for SSL in this work. However,
the above-mentioned network-based approach focuses on the optimization of the
label propagation result and pays little attention to the detailed dynamics of
the learning process itself. On the other hand, it is well-known that collective neural
dynamics generate rich information, and such a redundant processing
handles the adaptability and robustness of the learning process.  Moreover,
traditional graph-based techniques have high computational complexity, usually
at cubic order~\cite{Zhu2009a}.  A common strategy to overcome this disadvantage
is using a set of sparse prototypes derived from the data~\cite{Zhang2015}.
However, such a sampling process usually loses information of the original data.

Taking into account the facts above, we study a new type of dynamical
competitive learning mechanism in a complex network, called \textsl{particle
competition}. Consider a network where several particles walk and compete to
occupy as many vertices as possible while attempting to reject rival
particles. Each particle performs a combined random and preferential walk by
choosing a neighbor vertex to visit. Finally, it is expected that each particle
occupies a subset of vertices, called a community of the network. In this way,
community detection is a direct result of the particle competition. The
particle competition model was originally proposed in \cite{Quiles2008} and
extended for the data clustering task in~\cite{Silva2012}. Later, it has been
applied to semi-supervised learning~\cite{Thiago2012b,Thiago2012m} where the
particle competition is formally represented by a nonlinear stochastic dynamical
system. In all the models mentioned above, the authors concern vertex
dynamics---how each vertex changes its state (the level of dominance of each
particle). Intuitively, vertex dynamics is a rough modeling of a network because
each vertex can have several edges. A problem with the data analysis in this
approach is the overlapping nature of the
vertices, where a data item (a vertex in the networked form) can
belong to more than one class.  Therefore, it is interesting to
know how each edge changes its state in the competition process to acquire
detailed knowledge of the dynamical system.

In this paper, we propose a transductive semi-supervised learning model that
employs a vertex--edge dynamical system in complex networks. In this dynamical
system, namely
\ModelName, particles compete for \emph{edges} in a network.
Subnetworks are generated with the edges grouped by class dominance.  Here,
we call each subnetwork an \emph{unfolding}.
The learning model employs the unfoldings to
classify unlabeled data.  The proposed model offers satisfactory performance on
semi-supervised learning problems, in both artificial and real dataset.  Also,
it has shown to be suitable for detecting overlapping regions of data points by
simply counting the edges dominated by each class of particles.  Moreover, it
has low computational complexity order.

In comparison to the original particle competition models and other graph-based
semi-supervised learning techniques, the proposed one presents the following
salient features:

\paragraph{Particle competition dynamics occurs on nodes as well as on edges}
The inclusion of the edge domination model can give us more detailed information to
capture connectivity pattern of the input data. This is because there are much
more edges than vertices even in a sparse network. Consequently, the proposed
model has the benefit of granting essential information concerning overlapping
vertices.  Computer simulations show the proposed technique achieves a good
classification accuracy and it is suitable for situations with a small number of
labeled samples.

\paragraph{In the proposed model, particles are continuously generated and
removed from the system} Such a feature contrasts to previous particle
competition models that incorporate a preferential walking mechanism where
particles tend to avoid rival particles. As a consequence, the number of active
particles in the system varies over time. It is worth noting that the
elimination of preferential walking mechanism largely simplifies the dynamical
rules of particle competition model. Now, the new model is characterized by
the competition of only random walking particles, which, in turn, permits us to find
out an equivalent deterministic version. The original particle competition model
is intrinsically stochastic.  Then, each run may generate a different result.
Consequently, it has high computational cost. In this work, we find out a
deterministic system with running time independent of the number of particles,
and we demonstrate that it is mathematically equivalent to the stochastic model.
Moreover, the deterministic model has linear time order and ensures stable
learning. In other words, the model generates the same output for each run
with the same input. Furthermore, the system is simpler and easier to be
understood and implemented.  Thus, the proposed model is more
efficient than the original particle competition model.

\paragraph{There is no explicit objective function} In classical graph-based
semi-supervised learning techniques, usually, an objective function is defined
for optimization. Such function considers not only the label information, but
also the semi-supervised assumptions of smoothness, cluster,
or manifold. In particle competition models, we do not need to define an objective function.
Instead, dynamical rules which govern the time evolution of particles
and vertices (or edges) are defined.  Those dynamical rules mimic the phenomena
observed in some natural and social systems, such as resource competition among
animals, territory exploration by humans (or animals), election campaigns, etc. In
other words, the particle competition technique is typically inspired by nature.  In
such kind of technique, we have focused on behavior modeling instead of
objective modeling. Certain objectives can be achieved if the corresponding
behavioral rules are properly defined. In this way, we may classify classical
graph-based semi-supervised learning techniques as \emph{objective-based design} and
the particle competition technique as \emph{behavior-based design}.

The remainder of this paper is organized as follows. The proposed particle
competition system is studied in \cref{sec:system}. Our transductive
semi-supervised learning model is represented in \cref{sec:model}.  In
\cref{sec:simulation}, results of computer simulations are shown to
assess the proposed model performance on both artificial and real-world
datasets. Finally, \cref{sec:conclusion} concludes this paper.

\section{\ModelName}
\label{sec:system}

In this section, we give an introduction to the \ModelNameFirst---a particle competition system
for edge domination---%
explaining its basic design.  Whenever pertinent, we go into detail for further
clarification.


\subsection{Overview}

We consider a complex network expressed by a simple, unweighted, undirected
graph $G = (\VertexSet, \EdgeSet)$, where $\VertexSet$ is the set of vertices
and $\EdgeSet \subseteq \VertexSet \times \VertexSet$ is the set of edges.
If two vertices are considered similar, an edge connects them.
The network contains $\card{\VertexSet} = l + u$ vertices that can be either
labeled or unlabeled data points.
The set $\LabeledSet = \{v_{1}, \dots, v_{l}\}$ contains the labeled vertices,
where a vertex $v_i \in \LabeledSet$ has a label $y_i \in \{1, \ldots, C\}$.
We also use the terms \emph{label} and \emph{class} synonymously---if a vertex
is labeled with $c$, we say this vertex belongs to \class{c}.
The set $\UnlabeledSet = \{v_{l + 1}, \dots, v_{l + u}\}$ contains the unlabeled
vertices.
We suppose that $l \ll u$.
Thus, we have that $\LabeledSet\,\cap\,\UnlabeledSet = \emptyset$ and
$\VertexSet = \LabeledSet\,\cup\,\UnlabeledSet$.
The network is represented by the adjacency matrix $A = (a_{ij})$ where
$a_{ij} = a_{ji} = 1$ if \vertex{i} is connected to \vertex{j}.
We denote \edgenotation{i}{j} as the edge between vertices
\betweenvertices{i}{j}.
For practical reasons, we consider a connected network, and there is at least one labeled
vertex of each class.

In this model, particles are objects that flow within the network while
carrying a label.  Labeled vertices are \termdef{sources} for particles of the
same class and \termdef{sinks} for particles of other classes.  After
a particle is released, it randomly walks the network.  There is equal probability
among adjacent vertices to be chosen as the next vertex to be visited by the
particle.  Consider that a particle
is in \vertex{i}, it decides to move to \vertex{j} with
probability
\[
    \frac{a_{ij}}{\Degree{v_i}}
\]
with $\Degree{v_i}$ denoting the degree of \vertex{i}.

In each step, at the moment that a particle decides to move to a next vertex, it
can be absorbed (removed from the system).
If a particle is not
absorbed, we say that it has survived and it
remains active; and if it survives, then it continues walking.  Otherwise, the
particle is absorbed and ceases to affect the system.
The absorption depends on
the level of subordination and domination of a class against
all other classes in the edges.

To determine the level of domination and subordination of each class in
an edge, we take into account the active particles in the system.  The \termdef{current
directed domination} $\CurrentDomination{c}(t)$ is the number of active
particles belonging to \class{c} that decided to move from \vertex{i} to
\vertex{j} at \time{t} and survived.  Similarly, the
\termdef{current relative subordination} $\RelativeCurrentSubmission$
is the fraction of active particles that do not belong to \class{c} and have
successfully passed through \edge{i}{j}, \emph{regardless of direction}, at time
$t$.  Mathematically, we define the latter as
\begin{displaymath}
    \RelativeCurrentSubmission \defeq \begin{dcases*}
        1 - \frac{
            \CurrentDomination{c} + \CurrentDomination[ji]{c}
        }{
            \sum_{q=1}^C{
                \CurrentDomination{q} + \CurrentDomination[ji]{q}
            }
        } & if $\sum_{q=1}^C{\CurrentDomination{q} + \CurrentDomination[ji]{q}} > 0$, \\
        1-\frac{1}{C} & otherwise.
    \end{dcases*}
\end{displaymath}

The survival of a particle depends on the current relative subordination of the
edge and the destination vertex.  If a particle decides to move into a sink,
it will be absorbed with probability 1.  If the destination vertex is not a sink, its
survival probability is
\[
    1 - \SurvivalParam\RelativeCurrentSubmission(t)
\]
where $\SurvivalParam \in [0, 1]$ is a parameter for characterizing the competition level.

A source generates particles according to its degree and the current number of
active particles in the system.  Let $\onpart[](t)$ be the number of active particles belonging
to \class{c} in the system at \time{t}, a source generates new particles if
$\onpart[](t) < \onpart[](0)$.

Let $\SourceSet = \left\{v_i \middle\suchthat v_i \in \LabeledSet \mbox{ and } y_i = c
\right\}$ be the set of sources for particles that belong to \class{c},  the
number of newly generated particles belonging to \class{c} in \vertex{i} at \time{t} follows
the distribution
\[
    \begin{dcases*}
        \mathrm{B}\!\left( \onpart[](0)-\onpart[](t),~\Sourceness\right) &
        if $\onpart[](0)-\onpart[](t) > 0$, \\
        \mathrm{B}(1, 0) & otherwise,
    \end{dcases*}
\]
where
\[
    \Sourceness \defeq \begin{dcases*}
        \frac{
            \Degree{v_i}
        }{
            \sum_{v_j \in \SourceSet[j]}{
                \Degree{v_j}
            }
        } & if $v_i \in \SourceSet$, \\
        0 & otherwise,
    \end{dcases*}
\]
and $\mathrm{B}(n,p)$ is a binomial distribution.  In other words, if the number
of active particles is fewer than the initial number of particles, $\onpart[](0)$, each source performs
$\onpart[](0) - \onpart[](t)$ trials with probability $\Sourceness$ of
generating a new particle.

Therefore, the expected number of new particles belonging to \class{c} in
\vertex{i} at \time{t} is
\[
    \begin{dcases}
        \Sourceness\left(\onpart[](0)-\onpart[](t)\right) & \mbox{if } \onpart[](0)-\onpart[](t) > 0\mbox{,} \\
        0 & \mbox{otherwise.}
    \end{dcases}
\]

We are interested in the total number of visits of particles of each class to each edge.
Thus, we introduce the \emph{\DominationName}
$\oDomination{c}(t)$ that is the total number of particles belonging to
\class{c} that passed through \edge{i}{j} up to \time{t}.
Mathematically, this is defined as
\begin{equation}
  \label{eq:odomination}
    \oDomination{c}(t) \defeq \sum_{\tau=1}^t{
        \CurrentDomination{c}(\tau)\,\mbox{.}
    }
\end{equation}

Using the cumulative domination, we can group the edges by class domination.
For each \class{c}, the subset $\EdgeSet^c(t) \subseteq \EdgeSet$ is
\[
    \EdgeSet^c(t) \defeq \left\{
        \edgenotation{i}{j} \middle\suchthat \argmax_q\!{
            \left(
                \oDomination{q}(t) + \oDomination[ji]{q}(t)
            \right) = c
        }
    \right\}\mbox{.}
\]

We define the subnetwork
\begin{equation}
    \label{eq:unfolding}
    G^c(t) \defeq \left(\VertexSet, \EdgeSet^c(t)\right)
\end{equation}
as the \termdef{unfolding} of network $G$ according to \class{c} at
\time{t}.  We interpret the unfolding as a subspace with the most relevant
relationships for a given class. We use the available information in these
subnetworks for the study of overlapping regions and for semi-supervised
learning.


\subsection{An Illustrative Example}

\begin{figure}[!t]
  \centering
  \includegraphics[width=0.8\columnwidth]{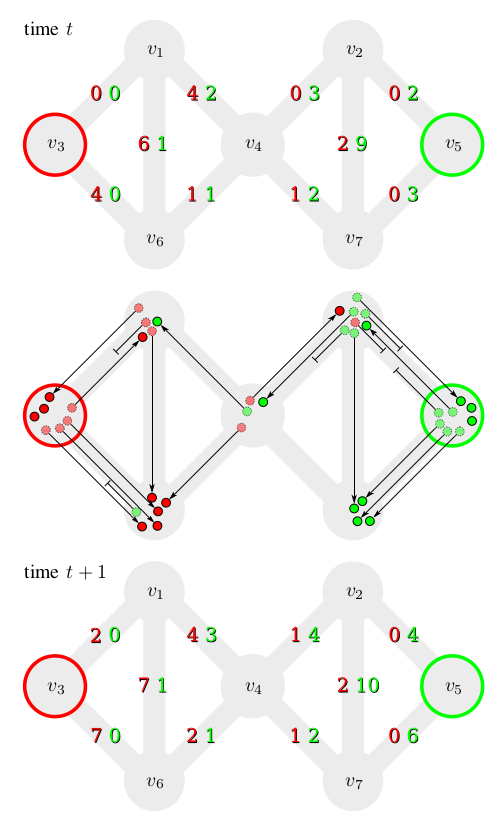}
  \caption{
    Illustration of one iteration of the system's evolution.  The network
    consists of 7 vertices and 10 edges; each color represents a
    label of a particle or a source.
    The first and the third networks depict the cumulative domination before and
    after the iteration.
    The cumulative domination is the number of visits of
    particles to an edge since the initial state.
    In the second network, particles are depicted in small circles.
    Active particles at \time{t} are depicted in
    dashed borders, whereas active particles at \time{t+1} are in full borders.
    An arrow indicates a particle movement, while an interrupted line indicates
    that the particle has been absorbed when trying to move through an edge.
    Particles without an adjacent arrow are generated by the sources at
    \time{t+1}.
  }
  \label{fig:example}
\end{figure}

One iteration of the system's evolution is illustrated by \cref{fig:example}.
The considered system contains 22 active particles at \time{t} and 20 at
\time{t+1}.  In an iteration, each particle moves to a neighbor vertex, without
preference. The movement of a particle is indicated by an arrow.  An interrupted
line indicates an edge in which the coming particle is absorbed.  A total of 6
particles are absorbed during this iteration, and the sources have
generated 4 new particles.

At \time{t}, for example, one of the red particles passing through \edge{1}{3} is
absorbed due to a current edge dominance of 0.5 in that edge (one red
particle and one green particle).  Conversely, all green particles that moved
through \edge{5}{7} remain active at \time{t+1}.  Since there is no rival
particle (red particle) passing through this edge, the updated value of the current
edge dominance is 1 and 0 for green and red classes, respectively.

In \edge{2}{4}, one red and two green particles chose to pass through.
One green particle is absorbed without affecting the new current level of
dominance.  Since one particle of each class successfully passed through
\edge{2}{4}, the new current level of dominance on this edge is 0.5.  The same
occurs for \edge{4}{7} where no particles have passed through and, thus, the
current level of dominance is set equally among all classes.

In \edges{2}{5} {3}{6}, particles have tried to move into a source of rival
particles (sinks). These particles are absorbed independently from the current
level of dominance.

Our edge-centric system can measure the overlapping
nature of the \vertex{4} by counting the edges dominated by each class, while a
vertex-centric approach would have lost such information.

\subsection{Mathematical Modeling}

Formally, we define the \ModelName as a dynamical \system{\tilde{X}(t)}.
The state of the system is
\begin{equation}
    \label{eq:osystem}
    \tilde{X}(t) \defeq \begin{bmatrix*}
        \ovpart(t) = \big[\onpart[i](t)\big]_i \\[0.1in]
        \oDMatrix(t) = \big(\oDomination{c}(t)\big)_{i,j}
    \end{bmatrix*}\,\mbox{,}
\end{equation}
where $\ovpart(t)$ is a vector, and each element
$\onpart[i](t)$ is the number of active particles belonging to \class{c} in
\vertex{i} at \time{t}.  Furthermore, $\oDMatrix(t)$ is a matrix whose elements
$\oDomination{c}(t)$ are given by \cref{eq:odomination}.

Let $\onewparticles(t)$ and $\tilde{a}^c_i(t)$ be,
respectively, the number of particles generated and absorbed by \vertex{i} at
\time{t}.  The evolution function~$\tilde\phi$ of the dynamical system is
\begin{displaymath}
    \tilde\phi: \begin{dcases}
        \onpart[i](t+1) =
        \onpart[i](t) + \sum_{j}{
            \left(\onpart[ji](t+1) - \onpart(t+1)\right)
        } \\
        \phantom{\onpart[i](t+1) = \onpart[i](t)} +
            \onewparticles(t+1) - \tilde{a}_i^c(t+1)\,\mbox{.}
            \phantom{\sum_j{}} \\
        \oDomination{c}(t+1) =
        \oDomination{c}(t) + \onpart(t+1)\,\mbox{.}
    \end{dcases}
\end{displaymath}

Intuitively, the number $\onpart[i]$ of active particles that are in a vertex is
the total number of particles arriving, $\onpart[ji]$, minus the number of
particles leaving, $\onpart[ij]$, or
being absorbed, $\tilde{a}_i^c$; additionally for labeled vertices, the
number of generated particles, $\onewparticles$.  Moreover, to calculate the total number
$\oDomination{c}$ of visits of particles to an edge, we simply add up the number
$\onpart$ at each time.  Values $\onpart$, $\onewparticles$, and $\tilde{a}^c_i$
are obtained \emph{stochastically} according to the dynamics of \WDynamicsName,
absorption, and generation.

The initial state of the system is given by an arbitrary number
$\onpart[i](0)$ of initial active particles and
\[
    \begin{dcases}
        \onpart[ij](0) = 0\,\mbox{,} \\
        \oDomination{c}(0) = 0\,\mbox{.}
    \end{dcases}
\]

To achieve the desirable network unfolding, it is necessary to average
the results of several simulations of the system with a very large number of
initial particles $\onpart[i](0)$.  Thus, the computational cost of such a
simulation is very high.  Conversely, we provide an alternative \system{X(t)}
that achieves similar results in a \emph{deterministic} manner.
More details will follow.

\subsection{Alternative Mathematical Modeling}

Consider the dynamical system
\begin{equation}
  \label{eq:system}
    X(t) \defeq \begin{bmatrix*}
        \vpart(t) = \big[\npart[i](t)\big]_i \\[0.1in]
        \Npart(t) = \big(\npart[ij](t)\big)_{i,j} \\[0.1in]
        \DMatrix(t) = \big(\Domination{c}(t)\big)_{i,j}
    \end{bmatrix*}\,\mbox{,}
\end{equation}
where $\vpart(t)$ is a row vector whose elements $\npart[i](t)$
give the population of particles with label $c$ in each vertex \vertex{i} at time $t$.
These values are associated to the number of active particles $\onpart[i]$
of \system{\tilde{X}}.  The elements $\npart[ij](t)$ and $\Domination{c}(t)$
of the sparse matrices $\Npart(t)$ and $\DMatrix(t)$ are related to the
\emph{current directed domination}, $\onpart[ij]$, and the \emph{cumulative
domination}, $\oDomination{c}$, respectively.  In other words, $\npart[ij](t)$
gives the number of particles of class $c$ that moved from \vertex{i} to
\vertex{j} at time $t$, while $\Domination{c}(t)$ gives the total number up to
time $t$.

The \system{X} is a nonlinear Markovian dynamical system with the deterministic
evolution function
\begin{equation}
    \label{eq:evolution}
    \phi\colon \begin{dcases}
        \vpart(t+1) = \vpart(t) \times \FinalTrans(X(t)) + \vnewparticles(X(t))\\
        \Npart(t+1) = \diag{\vpart(t)} \times \FinalTrans(X(t)) \\
        \DMatrix(t+1)  = \DMatrix(t) + \Npart(t+1)\,\mbox{,}
    \end{dcases}
\end{equation}
where $\diag{\vec{v}}$ is a square matrix with the elements of vector $\vec{v}$
on the main diagonal and $\times$ stands for the vector-matrix product.

The function $\FinalTrans(X(t))$ of the \system{X} at time $t$ gives a square
matrix whose elements are
\begin{equation}
    \label{eq:prob}
    \resizebox{\columnwidth}{!}{$%
        \finaltrans(X(t)) \defeq \begin{dcases*}
            0 & if $v_j \in \LabeledSet$ and $y_j \neq c\,$, \\
            \frac{a_{ij}}{\Degree{v_i}}\left(
                1 - \SurvivalParam\submission(X(t))
            \right) & otherwise,
        \end{dcases*}%
    $}
\end{equation}
where
\begin{multline}
    \label{eq:subordination}
    \submission\!\left(X(t)\right) \defeq \begin{dcases*}
    1 - \frac{
        \xnpart[ij]{c}(t) + \xnpart[ji]{c}(t)
    }{
        S
    } & if  $S > 0$,\\
    1-\frac{1}{C} & otherwise,
    \end{dcases*} \\
    \mbox{with } S = \sum_{q=1}^C{
            \xnpart[ij]{q}(t) + \xnpart[ji]{q}(t)
        }\mbox{.}
\end{multline}

Given that we know the initial state $X(0)$ of the system, the function
$\vnewparticles(X(t))$ of the \system{X} at time $t$ returns a row vector
where the $i$-th element is
\begin{equation}
    \label{eq:gen}
    \newparticles(X(t)) \defeq
    \Sourceness\MaxSet{0}{
        \vec{1}\cdot\vpart(0)-\vec{1}\cdot\vpart(t)
    }\mbox{,}
\end{equation}
where $\vec{1}$ is a row vector whose elements are $1$, and $\cdot$ stands
for the inner product between vectors.

The initial state of the \system{X} is given by an arbitrary \emph{population
size}\footnote{%
  In \system{X}, vector $\vpart(t)$ describes the quantity of
  particles in each vertex.  Since $X$ has multiplicative scaling behavior,
  $\vpart(t)$ is not necessarily composed only of integer values; $\vpart(t)$
  values can be a discrete distribution of particles.  See \cref{sub:scale} for
  more details.%
} $\npart[i](0)$ of initial active particles and
\[
    \begin{dcases}
        \npart[ij](0) = 0\,\mbox{,} \\
        \Domination{c}(0) = 0\,\mbox{.}
    \end{dcases}
\]


If the initial number of particles in each vertex in \system{\tilde{X}}
is proportional to the initial population size in \system{X}, we
provide evidence that the unfolding result tends to be the same for both
systems---represented by \cref{eq:osystem} and \cref{eq:system}, respectively---, as $\onpart[i](0) \rightarrow \infty$ for all $c \in \{1, \dots, C\}$
and $i \in \{1, \dots, \card{\VertexSet}\}$.


\subsection{Mathematical Analysis}
\label{sub:analysis}

In the previous subsections, we modeled two possibly equivalent systems,
\systemnotation{X} and~\systemnotation{\tilde{X}}.  In this section, we
present mathematical results that prove the equivalence of the two systems
under certain assumptions.

\begin{theorem}{Systems $X$ and $\tilde{X}$ are asymptotically equivalent if the
following conditions hold:}
    \label{thm:eq}
    \[
        \expect{\RelativeCurrentSubmission(t)} \rightarrow
        1 - \frac{
            \expect{\CurrentDomination{c}(t)} + \expect{\CurrentDomination[ji]{c}(t)}
        }{
            \sum_{q=1}^C{
                \expect{\CurrentDomination{q}(t)} + \expect{\CurrentDomination[ji]{q}(t)}
            }
        }\mbox{ and}
    \]
    \[
        \expect{\onewparticles(t+1)} \rightarrow
        \Sourceness\MaxSet{0}{
            \onpart[](0)-\expect{\onpart[](t)}
        }\mbox{ as}
    \]
    \[
        \onpart[i](0) \rightarrow \infty\mbox{,}
    \]
    for all $i,j \in \VertexSet$, $t > 0$, and $c \in \{1, \dots, C\}$,
    we have
    \[
        \npart[i](t) = \kappa\expect{\onpart[i](t)}\mbox{, }
        \npart(t) = \kappa\expect{\onpart(t)}\mbox{, and}
    \]
    \[
        \Domination{c}(t) = \kappa\expect{\oDomination{c}(t)}\mbox{, }
    \]
    for some $k > 0$ constant.
\end{theorem}

In order to prove \cref{thm:eq}, we study the following mechanisms of the
particle competition system:

\subsubsection{Particle motion and absorption}

In the proposed system, each particle moves independently from the others.
Particle's movement through an edge affects the absorption of rival particles only
in the next iteration.  Such conditions are favorable to naturally regard the
system's evolution in terms of the distribution of particles over the network.
Next, we present a formal model for particle movement.

Let $\ParticleMoved{p}$ be a discrete random variable that is~1 if \particle{p}
was in \vertex{i} at \time{t} and moved into \vertex{j} at \time{t + 1};
and it is~0 otherwise.  Since each particle in a vertex moves independently, we
can write this probability in terms of a particle's class; that is,
$\AParticleMoved(t+1) = \ParticleMoved{p}$ for any \particle{p} that belongs to
\class{c} and is in \vertex{i} at \time{t}.

The probability $\ofinaltrans{t+1}$ is affected by the movement decision of
a particle and whether it was absorbed after the decision.
By formulation, in dynamical \system{\tilde{X}} the conditional
probability, given that $\RelativeCurrentSubmission(t) = \xi$, is
\begin{multline*}
    \cofinaltrans{\xi} \\
    = \begin{dcases*}
        0 & if $v_j \in \LabeledSet$ and $y_j \neq c\,$, \\
        \frac{a_{ij}}{\Degree{v_i}}\left(
            1 - \SurvivalParam\cdot\xi
        \right) & otherwise.
    \end{dcases*}
\end{multline*}

That is, when a particle tries to move into a sink, the survival probability
is zero.  Otherwise, a particle only reaches \vertex{j} if it chooses to
move into the vertex and it is not absorbed.

Let $f_{\RelativeCurrentSubmission(t)}$ be the probability density function of
the random variable $\RelativeCurrentSubmission$.  Hence, the probability
$\ofinaltrans{t + 1}$ is
\begin{multline*}
    \int_{-\infty}^{\infty}{
        \cofinaltrans{\xi}f_{\RelativeCurrentSubmission(t)}(\xi)\dd\xi
    }\\
    = \int_{-\infty}^{\infty}{
        \frac{a_{ij}}{\Degree{v_i}}\left(
            1 - \SurvivalParam\cdot\xi
        \right)f_{\RelativeCurrentSubmission(t)}(\xi)\dd\xi
    }\\
    = \frac{a_{ij}}{\Degree{v_i}}\left(
        \int_{-\infty}^{\infty}{
            f_{\RelativeCurrentSubmission(t)}(\xi)\dd\xi
        }
        -
        \SurvivalParam\int_{-\infty}^{\infty}{
            \xi f_{\RelativeCurrentSubmission(t)}(\xi)\dd\xi
        }
    \right)\\
    = \frac{a_{ij}}{\Degree{v_i}}\left(
        1 - \SurvivalParam\expect{\RelativeCurrentSubmission(t)}
    \right)\mbox{,}
\end{multline*}
if $v_j \in \UnlabeledSet$ or $v_j \in \LabeledSet \land y_j \neq c\,$. Otherwise, it is zero.

Furthermore, $\RelativeCurrentSubmission$ is convex with fixed values of
$\CurrentDomination[ij]{q}, \CurrentDomination[ji]{q}$ for all $q\neq c$.
Thus, with the Jensen's inequality~\cite{Jensen1906}, we have
\begin{equation}
    \label{eq:submission}
    \expect{\RelativeCurrentSubmission(t)} \geq
    1 - \frac{
        \expect{\CurrentDomination{c}(t)} + \expect{\CurrentDomination[ji]{c}(t)}
    }{
        \sum_{q=1}^C{
            \expect{\CurrentDomination{q}(t)} + \expect{\CurrentDomination[ji]{q}(t)}
        }
    }\mbox{.}
\end{equation}

\subsubsection{Particle generation}

In dynamical \system{\tilde{X}} the expected number of particles belonging to
\class{c} generated at \vertex{i} at \time{t+1} is
\[
    \expect{\onewparticles(t+1)} = \sum_{\eta=0}^\infty{
        \cexpect{\onewparticles(t+1)}{\onpart[](t) = \eta}
        \prob{\onpart[](t) = \eta}
    }\,\mbox{.}
\]

The conditional expectation $\cexpect{\onewparticles(t+1)}{\onpart[](t) = \eta}$
is, by formulation,
\[
    \Sourceness\cdot\MaxSet{0}{
        \onpart[](0)-\eta
    }\mbox{,}
\]
and thus, $\expect{\onewparticles(t+1)}$ is%
\newcommand{\tmp}{\onpart[](0)-\onpart[](t)}%
\begin{multline*}
    \Sourceness\sum_{\eta=0}^\infty{
        \MaxSet{0}{
            \onpart[](0)-\eta
        }
        \prob{\onpart[](t) = \eta}
    }\\
    = \Sourceness\,\expect{\MaxSet{0}{\tmp}}\mbox{.}
\end{multline*}

Since $\MaxSet{0}{x}$ is convex for all $x\in\mathbb{R}$ and according
to Jensen's inequality, we have%
\renewcommand{\tmp}{\onpart[](0)-\expect{\onpart[](t)}}%
\begin{equation}
    \label{eq:generation}
    \expect{\onewparticles(t+1)} \geq \Sourceness\MaxSet{0}{\tmp}\mbox{.}
\end{equation}

\subsubsection{Expected edge domination}

At the beginning of \system{\tilde{X}} we have
\[
    \oDomination{c}(0) = 0
\]
and, for $t \geq 0$,
\begin{multline}
    \label{eq:oDomination}
    \expect{\oDomination{c}(t+1)} = \expect{\oDomination{c}(t) +
    \onpart(t+1)} \\
    = \expect{\oDomination{c}(t)} + \expect{\onpart(t+1)}\mbox{.}
\end{multline}

Given that $\onpart[i](t) = \eta$ is known and
since each particle in a vertex moves independently,
the number of particles that successfully reaches \vertex{j} at
\time{t+1} is
\[
    \onpart(t+1) = \sum_{k=1}^{\eta}{
        \ParticleMoved{p_k}
    }\,\mbox{,}
\]
where $p_k$ is a particle that belongs to \class{c} and is in \vertex{i}.
Then, the expected value $\expect{\onpart(t+1)}$ is
\begin{multline*}
    \sum_{\eta=0}^\infty{
        \cexpect{\onpart(t+1)}{\onpart[i](t) = \eta}
        \prob{\onpart[i](t) = \eta}
    } \\
    = \sum_{\eta=0}^\infty{
        \prob{\onpart[i](t) = \eta}\cdot\sum_{k=1}^{\eta}{
            \expect{\ParticleMoved{p_k}}
        }
    }\\
    = \sum_{\eta=0}^\infty{
        \prob{\onpart[i](t) = \eta}\cdot\sum_{k=1}^{\eta}{
            \prob{\ParticleMoved{p_k} = 1}
        }
    }\\
    = \sum_{\eta=0}^\infty{
        \eta\prob{\onpart[i](t) = \eta}\cdot\ofinaltrans{t}
    }\,\mbox{.}
\end{multline*}
Finally,
\begin{equation}
    \label{eq:onpart}
    \expect{\onpart(t+1)} = \expect{\onpart[i](t)}\ofinaltrans{t}
\end{equation}
for all $t \geq 0$, $c = \{1, \dots, C\}$, and
$i,j\in\{1,\dots,\card{\VertexSet}\}$.

\subsubsection{Expected number of particles}

We know the number of particles at the beginning of
\system{\tilde{X}(t)}, so
\[
    \expect{\onpart[i](0)} = \onpart[i](0)
\]
and, for all $t \geq 0$, the expected value $\expect{\onpart[i](t+1)}$ is
\begin{multline*}
    \expect{\onpart[i](t)} + \sum_{j}{\left(
        \expect{\onpart[ji](t+1)} - \expect{\onpart(t+1)}
    \right)}\\
    {} + \expect{\onewparticles(t+1)} -
        \expect{\tilde{a}^c_i(t+1)}\mbox{.}
\end{multline*}

However, the expected number of particles that were absorbed in \vertex{i} is
the expected number of particles in \vertex{i} minus the expected number of
particles that survived when moving away.  Thus, $\expect{\onpart[i](t+1)}$ can
be written as
\begin{multline*}
    \expect{\onpart[i](t)} +
    \sum_{j}{\left(
        \expect{\onpart[ji](t+1)} - \expect{\onpart(t+1)}
    \right)} \\
    {} + \expect{\onewparticles(t+1)} - \left(
        \expect{\onpart[i](t)} - \sum_{j}{
            \expect{\onpart(t+1)}
        }
    \right)\mbox{.}
\end{multline*}

And, finally
\begin{equation}
    \label{eq:onparti}
    \expect{\onpart[i](t+1)} = \sum_{j}{
        \expect{\onpart[ji](t+1)}
    } + \expect{\onewparticles(t+1)}\mbox{,}
\end{equation}
for all $t \geq 0$, $c = \{1, \dots, C\}$, and
$i\in\{1,\dots,\card{\VertexSet}\}$.

\subsubsection{Scale invariance}
\label{sub:scale}

The unfolding $G^c(t)$ from \system{X} is invariant under real
positive multiplication of the row vector $\vpart(0)$.  In order to prove this
property, consider the following lemma.

\begin{lemma}
    \label{lem:scale}
    System $X$ has positive multiplicative scaling behavior of order 1.
    Given an arbitrary initial state $X_0$ of the \system{X}, it means that
    \begin{multline}
        \label{eq:scale}
        X(t) = X_t~\suchthat~X(0) = X_0 \\ \iff X(t) = \kappa X_t~\suchthat~X(0) =
        \kappa X_0
    \end{multline}
    for all $t>0$ and $\kappa > 0$.
\end{lemma}

\begin{IEEEproof}[Proof of \cref{lem:scale}]
    First, we show that the functions $\FinalTrans$ are invariant to parameter
    scaling.  Given an arbitrary system state $X(t) = X_t$ and $\kappa > 0$,
    \begin{multline*}
        \finaltrans(\kappa X_{t}) = \finaltrans(X_{t}) \\
        = \begin{dcases*}
            0 & if $v_j \in \LabeledSet$ and $y_j \neq c\,$, \\
            \frac{a_{ij}}{\Degree{v_i}}\left(
                1 - \SurvivalParam\submission(X_{t})
            \right) & otherwise,
        \end{dcases*}
    \end{multline*}
    since the term $\submission$ can be either
    \begin{multline*}
        \submission(\kappa X_t) = 1 - \frac{
            \kappa \xnpart[ij]{c}(t; X_t) + \kappa \xnpart[ji]{c}(t; X_t)
        }{
            \sum_{q=1}^C{
                \kappa \xnpart[ij]{q}(t; X_t) + \kappa \xnpart[ji]{q}(t; X_t)
            }
        } \\
        = 1 - \frac{
            \xnpart[ij]{c}(t; X_t) + \xnpart[ji]{c}(t; X_t)
        }{
            \sum_{q=1}^C{
                \xnpart[ij]{q}(t; X_t) + \xnpart[ji]{q}(t; X_t)
            }
        }
        = \submission(X_t)
    \end{multline*}
    or
    \[
        \submission(\kappa X_{t}) = \submission(X_{t}) =
        1-\frac{1}{C}\text{.}
    \]

    Now, consider two arbitrary initial states,
    \[
        X_0 = \begin{dcases}
            \npart[i](0) = \eta_i \\
            \npart[ij](0) = 0 \\
            \Domination{c}(0) = 0
        \end{dcases} \qquad\mbox{and}\qquad
        \kappa X_0 = \begin{dcases}
            \npart[i](0) = \kappa \eta_i \\
            \npart[ij](0) = 0 \\
            \Domination{c}(0) = 0
        \end{dcases}\mbox{,}
    \]
    for all $i, j, c$.  We have that,
    \begin{multline*}
        \npart(1; \kappa X_0)
        = \npart(0; \kappa X_0) \finaltrans(\kappa X_0)\\
        = \kappa \npart(0; X_0) \finaltrans(X_0)
        = \kappa \npart(1; X_0)\,\mbox{,}
    \end{multline*}
    \begin{multline*}
        \npart[i](1; \kappa X_0)
        = \sum_{j}{\npart[ji](1; \kappa X_0)} +
        \newparticles\!\left(\kappa X_0\right) \\
        = \kappa \sum_{j}{\npart[ji](1; X_0)} + 0
        = \kappa \npart[i](1; X_0)\mbox{,}%
    \end{multline*}
    and
    \begin{multline*}
        \Domination{c}(1; \kappa X_0) = \Domination{c}(0; \kappa X_0) +
        \npart(1; \kappa X_0) = \\
        0 + \kappa \npart(1; X_0) = \kappa \Domination{c}(1; X_0)
    \end{multline*}
    Thus, Relation~\eqref{eq:scale} holds true for $t = 1$.

    Assuming that Relation~\eqref{eq:scale} holds true for some time $t$, we
    show that the relation holds true for $t+1$:
    \begin{multline*}
        \npart(t+1; \kappa X_0)
        = \npart(t; \kappa X_0)\finaltrans(\kappa X_t) \\
        = \kappa \npart(t; X_0)\finaltrans(X_{t})
        = \kappa \npart(t+1; X_0)
    \end{multline*}
    \begin{multline*}
        \npart[i](t+1; \kappa X_0)
        = \sum_{j}{\npart[ji](t+1; \kappa X_0)} +
        \newparticles\!\left(\kappa X_t\right) \\
        = \kappa \sum_{j}{\npart[ji](t+1; X_0)} +
        \kappa\newparticles\!\left(X_t\right)
        = \kappa\npart[i](t+1; X_0)
    \end{multline*}
    since
    \[
        \newparticles(\kappa X_t)
        = \Sourceness\MaxSet{0}{
            \kappa\vec{1}\cdot\vpart(0; X_0)
            -\kappa\vec{1}\cdot\vpart(t; X_t)
        }\mbox{,}
    \]
    and
    \begin{multline*}
        \Domination{c}(t+1; \kappa X_0) = \Domination{c}(t; \kappa X_0) +
        \npart(t+1; \kappa X_0) \\
        = \kappa \Domination{c}(t; X_0) + \kappa\npart(t+1; X_0)
        = \kappa\Domination{c}(t+1; X_0)
    \end{multline*}
    So Relation~\eqref{eq:scale} indeed holds true for $t+1$.

    Since both the basis and the inductive step have been performed, by mathematical
    induction, the lemma is proved for all $t \geq 0$ natural.
\end{IEEEproof}

Finally, using these studies, we may prove the theorem.

\begin{IEEEproof}[Proof of~\cref{thm:eq}]
    By \cref{eq:oDomination,eq:onpart,eq:onparti}, we have
    \[
        \begin{dcases*}
            \expect{\onpart[i](t+1)} = \sum_{j}{
                \expect{\onpart[ji](t+1)}
            } + \expect{\onewparticles(t+1)}\mbox{,}\\
            \expect{\onpart(t+1)} = \expect{\onpart[i](t)}\ofinaltrans{t}\mbox{,}
            \phantom{\sum_j{}}\\
            \expect{\oDomination{c}(t+1)} =\expect{\oDomination{c}(t)} +
            \expect{\onpart(t+1)}\mbox{,}
        \end{dcases*}
    \]
    which is \system{X} assuming that Inequalities~\eqref{eq:submission}
    and~\eqref{eq:generation} tend to equality when there is a large number of
    particles and $\kappa\onpart[i](0) = \npart[i](0)$, for any $k>0$ constant (scale
    invariance property).
\end{IEEEproof}

\begin{remark}
    Even if the convergence of Inequalities~\eqref{eq:submission}
    and~\eqref{eq:generation} are not true, another property that possibly
    makes the two systems equivalent is the compensation over time.  At the
    beginning, both systems are equal;  however, in the next
    iteration both absorption probability~\eqref{eq:submission} and generated
    particles~\eqref{eq:generation} are underestimated.  Consequently, particles
    that have survived may compensate the ones that were not generated.
    Furthermore, the lower the number of absorbed particles in an iteration, the
    higher the absorption probability in the next iteration.  Likewise, the
    lower the number of generated particles in an iteration, the higher is the
    expected number of new particles in the next iteration.
\end{remark}


\section{Semi-supervised Learning by Labeled Component Unfolding}
\label{sec:model}


\newcommand{\TrainingSet}{\mathcal{X}_{\text{labeled}}}
\newcommand{\TestSet}{\mathcal{X}_{\text{unlabeled}}}

Unfoldings generated by \ShortModelName are incorpored in a
semi-supervised learning model.  Consider two sets
$\TrainingSet = \{x_{1}, \dots, x_{l}\}$ and
$\TestSet = \{x_{l+1}, \dots, x_{l+u}\}$
such that $x_i \in \RealSet^D$ for all $i$.
Each data point $x_{i} \in \TrainingSet$ is associated to a label
$y_i \in \{1, \dots, C\}$.
In the semi-supervised learning setting, our goal is to correctly assign
existing labels to the unlabeled data $\TestSet$.

In short, the proposed learning model has three steps:
\begin{inparaenum}[\itshape a\upshape)]
\item a network is constructed based on a dataset composed of feature vectors,
  where vertices represent
    data points, and edges represent similarity relationship;
\item \ShortModelName is applied to obtain the unfoldings, that is, a distinct set of
    edges for each class of the dataset; and
\item infer labels for every data point in $\TestSet$.
\end{inparaenum}

Next, each step of the proposed learning model is presented in detail.
Further to the model's algorithm description, its computational complexity
analysis is also presented.


Since the proposed dynamical system takes place on a complex network, the
original dataset needs to be represented in a network structure.  Therefore,
the first step of our learning model is to obtain a network representation.
Each data point is
associated to a single vertex of the network.  Moreover, the network must
be sparse, undirected, and unweighted.  Labeled vertices
correspond to the set of points in $\TrainingSet$, and unlabeled vertices
to the set of points in $\TestSet$.  Two vertices are connected by an edge if
they have a relationship of similarity, which is determined by some metric
or by the particular problem.
Any graph construction method that satisfies such conditions may be used in this
step.  The \emph{k}-Nearest Neighbor (\kNN) graph construction method is one of them.

The second step is to run \system{X} defined by \cref{eq:evolution} using the constructed complex network
as its input.
Two conditions are satisfied on the system initialization.  First, no class should
be privileged. Second, during the first iterations, all particles should be able
to flow within the network with a small probability of absorption.  Thus, the
initial conditions of the system, for all $i$, $j$, and $c \in \{1, \dots, C\}$,
are
\begin{equation}
    \label{eq:initial}
    \begin{split}
        \npart[i](0) &= \frac{\Degree{v_i}}{2 \card{\EdgeSet}}\mbox{, }\\
        \npart(0) &= 0\,\mbox{,}\\
        \Domination{c}(0) &= 0\,\mbox{.}
    \end{split}
\end{equation}

Since there are always particles in the system, the iteration of \system{X} should be stopped
if the time limit has been reached.  The time limit parameter $\tau$ controls
the maximum number of iterations of the system.

At the last step, the networks $G^c(\tau)$ are used for vertex
classification.  We assign a label $y_j \in \{1, \dots, C\}$ for each unlabeled
vertex $v_j \in \UnlabeledSet$, with the information provided by the networks
$G^c$.  Label $y_j$ is assigned based on the density of edges in its
neighborhood.  Formally, the label for \vertex{j} is
\begin{equation}
    \label{eq:labeling}
    y_j = \argmax_{c\in\{1,\dots,C\}}\!{
        \card{\EdgeSet(\mathcal{N}_{c,j})}
    }\,\mbox{,}
\end{equation}
where $\mathcal{N}_{c,j}$ is the neighborhood of \vertex{j}
in the unfolding $G^c(\tau)$.  We denote the number of edges in this neighborhood
as $\card{\EdgeSet(\mathcal{N}_{c,j})}$.


\subsection{Algorithm}

\newcommand{\MeanDeg}{\langle k \rangle}

\begin{algorithm}[t]
\caption{Semi-supervised Learning by LCU.}
\label{alg:learning}
\newcommand{\VarUnfoldResult}{subnetworks\xspace}
\begin{algorithmic}[1]
    \Function{Classifier}{$\TrainingSet$, $\TestSet$, $\SurvivalParam$,
    $\tau$}
        \State $G \gets$ \Call{BuildNetwork}{$\TrainingSet$, $\TestSet$}
        \State \VarUnfoldResult $\gets$ \Call{Unfold}{$G$, $\SurvivalParam$, $\tau$}
        \State \Return \Call{Classify}{$\TestSet$, \VarUnfoldResult}
    \EndFunction
\end{algorithmic}
\end{algorithm}

\begin{algorithm}[t]
\caption{\ModelName.}
\label{alg:unfold}
\begin{algorithmic}[1]
    \Function{Unfold}{$G$, $\SurvivalParam$, $\tau$}

        \For{$c \in \{1, \dots, C\}$}
            \State $\vpart \gets$ \Call{$n_0$}{$G$, $c$}
            \Comment{\cref{eq:initial}}
            \State $\Npart \gets$ \Call{$N_0$}{$G$, $c$}
            \State $\DMatrix \gets$ \Call{$\Delta_0$}{$G$, $c$}
        \EndFor

        \For{$t \in \{1, \dots, \tau\}$}

            \For{$c \in \{1, \dots, C\}$}
                \State $\FinalTrans \gets$ \Call{$P$}{%
                    $G$, $\Npart[1]$, \dots, $\Npart[C]$, $\SurvivalParam$%
                }
                \Comment{\cref{eq:prob}}
                \State $\vnewparticles \gets$ \Call{$g$}{%
                    $G$, $\vpart$, $t$%
                }
                \Comment{\cref{eq:gen}}
                \State $\Npart \gets \diag{\vpart}\times\FinalTrans$
                \State $\vpart \gets \vpart\times\FinalTrans + \vnewparticles$
                \State $\DMatrix \gets \DMatrix + \Npart$
            \EndFor

        \EndFor

        \State \Return \Call{Subnetworks}{$G$, $\DMatrix$}
        \Comment{\cref{eq:unfolding}}
    \EndFunction
\end{algorithmic}
\end{algorithm}

\cref{alg:learning} summarizes the steps of our learning model.  The algorithm
accepts the labeled dataset $\TrainingSet$, the unlabeled dataset $\TestSet$,
and 2 user-defined parameters---the competition ($\SurvivalParam$)
parameter of the \system{X} and the time limit parameter ($\tau$).
Moreover, it is necessary to choose a network formation technique.

The first step of the learning model is mapping the original vector-formed data to a network
using a chosen network formation technique. Afterward, we unfold the network as
described in \cref{alg:unfold}.  This algorithm iterates the \ShortModelName
to produce one subnetwork for each class.  Steps 2--6 initialize the system state
as indicated in \cref{eq:initial}.  Steps 7--15
iterate the system until $\tau$ using the evolution function~$\phi$
\eqref{eq:evolution}.  Step 16 calculates and returns the unfoldings for each
class.  Back to \cref{alg:learning}, by using the produced unfoldings, the
unlabeled data are classified as described in \cref{eq:labeling}.

\subsection{Computational Complexity and Running Time}

Here, we provide the computational complexity analysis step by step.

The construction of the complex network from the input dataset depends on the
chosen method.  Since $\card{\VertexSet} =
\card{\TrainingSet} + \card{\TestSet}$ is the number of data samples.  The \kNN
method, for example, has complexity order of $\Complexity{D
\card{\VertexSet}\log{\card{\VertexSet}}}$ using multidimensional binary search
tree~\cite{Bentley1975}.

The second step is running \system{X} defined by \cref{eq:evolution}.  Using sparse matrices, the system
initialization, steps 2--6 of \cref{alg:unfold}, has complexity order of
$\Complexity{C \card{\VertexSet} + C \card{\EdgeSet}}$.  The system iteration
calculates $\tau C$ times the evolution function~$\phi$ \eqref{eq:evolution}
represented in steps 8--14.  The time complexity of each part of the system
evolution is presented below.
\begin{itemize}
    \item Step 9, computation of the matrix $P^c$.  This matrix has
        $\card{\EdgeSet}$ non-zero entries.  It is necessary to calculate
        $\sigma^c_{ij}$ for each non-zero entry.  Hence, this step has
        complexity order of $\Complexity{C\card{\EdgeSet}}$.  However, the
        denominator of \cref{eq:subordination} is the same for all values of $c$.
    \item Step 10, computation of the vector $\vec{g}^c$.  This vector has
        $\card{\LabeledSet}$ non-zero entries.  It is also necessary to
        calculate the total number of particles in the system.  So, this calculation has time
        complexity order of $\Complexity{\card{\LabeledSet} + \card{\VertexSet}}$.
    \item Step 11, computation of the matrix $N^c$.  The multiplication between
        a diagonal matrix and a sparse matrix with $\card{\EdgeSet}$ non-zero
        entries has time complexity order of
        $\Complexity{\card{\EdgeSet}}$.
    \item Step 12, computation of the vector $\vec{n}^c$. Suppose that $\MeanDeg$
        is the average vertex degree of the input network; it follows that this
        can be performed in $\Complexity{\card{\VertexSet}\MeanDeg} =
        \Complexity{\card{\EdgeSet}}$.
    \item Step 13, computation of the matrix $\Delta^c$.  This sparse matrix
        summation has complexity order of $\Complexity{\card{\EdgeSet}}$.
\end{itemize}
After the system evolution, the unfolding process performs
$\Complexity{C\card{\EdgeSet}}$ operations.  Thus, the total time complexity
order of the system simulation is $\Complexity{\tau C \card{\EdgeSet} + \tau C
\card{\VertexSet}}$.  However,
the value of $\tau$ is fixed and the value of $C$ is usually very small.

The vertex labeling step is the last step of the learning model.  The
time complexity of this step depends on the calculation of the number of edges
in the neighborhood of each unlabeled vertex in each unfolding.
It can be efficiently calculated by using one step of a
breadth-first search in~$G^c$.  Hence, the order of the average-time complexity
is \mbox{$\Complexity{C \card{\UnlabeledSet} \MeanDeg^2} \approx
\Complexity{C \card{\EdgeSet}}$.
}

\begin{table}
    \centering
    \caption{
        Time Complexity of Common Graph-based Techniques Disregarding the Graph
        Construction Step
    }
    \label{tbl:time-comparison}
    \begin{tabular}{l c}
        \toprule
        Algorithm & Time Complexity \\
        \midrule
        Transductive SVM~\cite{Vapnik1998} & $C\card{\VertexSet}^3$ \\
        Local and Global Consistency~\cite{Zhou2004} & $\card{\VertexSet}^3$ \\
        Large Scale Transductive SVM~\cite{Collobert2006} & $C\card{\VertexSet}^2$ \\
        Dynamic Label Propagation~\cite{Wang2013} & $\card{\VertexSet}^2$ \\
        Label Propagation~\cite{Zhu2002} & $\card{\VertexSet}^2$ \\
        Original Particle Competition~\cite{Silva2012} & $C^2\card{\VertexSet} + C\card{\EdgeSet}$ \\
        \bf Labeled Component Unfolding & $C\card{\VertexSet} + C\card{\EdgeSet}$ \\
        \bf Minimum Tree Cut~\cite{Zhang2014a} & $\card{\VertexSet}$ \\
        \bottomrule
    \end{tabular}
\end{table}

In summary, considering the discussion above, our learning model runs in
$\Complexity{D\card{\VertexSet}\log{\card{\VertexSet}} + C\card{\EdgeSet} +
C\card{\VertexSet}}$ including the transformation from vector-based dataset to
a network.
    \cref{tbl:time-comparison} compares the time
    complexity of common graph-based techniques disregarding the graph
    construction step.  Only the proposed LCU method and Minimum Tree
    Cut~\cite{Zhang2014a} have linear time, though the latter must
    either receive or construct a spanning tree.  Consequently, the Minimum Tree Cut has
    a performance similar to the scalable version of traditional algorithms,
    such as those using subsampling practices.

\begin{figure}[!t]
    \centering
    \subfloat[]{
        \label{time-vertex}
        \includegraphics[width=0.9\columnwidth]{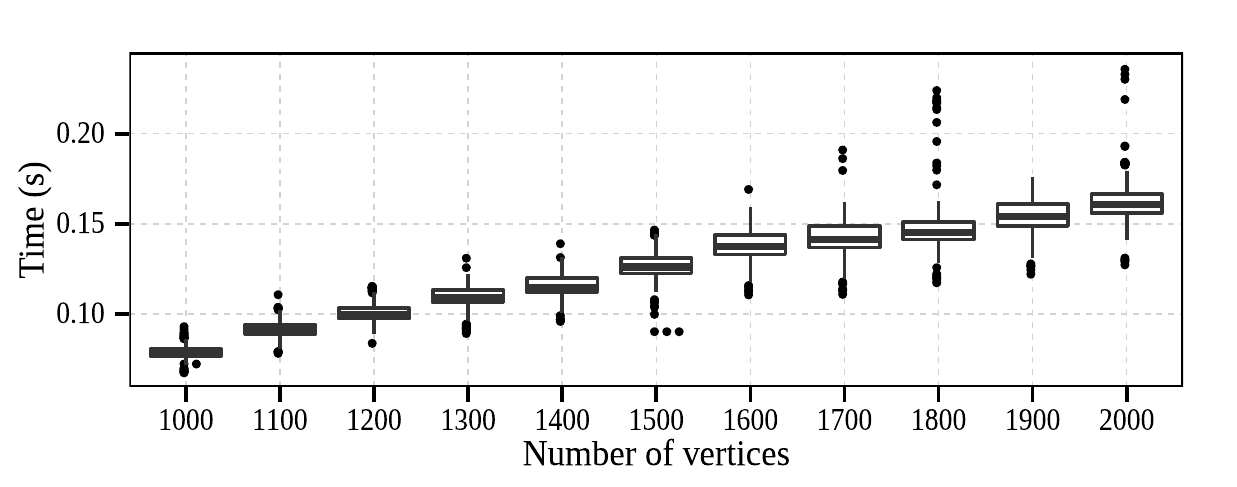}
    }
    \hfil
    \subfloat[]{
        \label{fig:time-edge}
        \includegraphics[width=0.9\columnwidth]{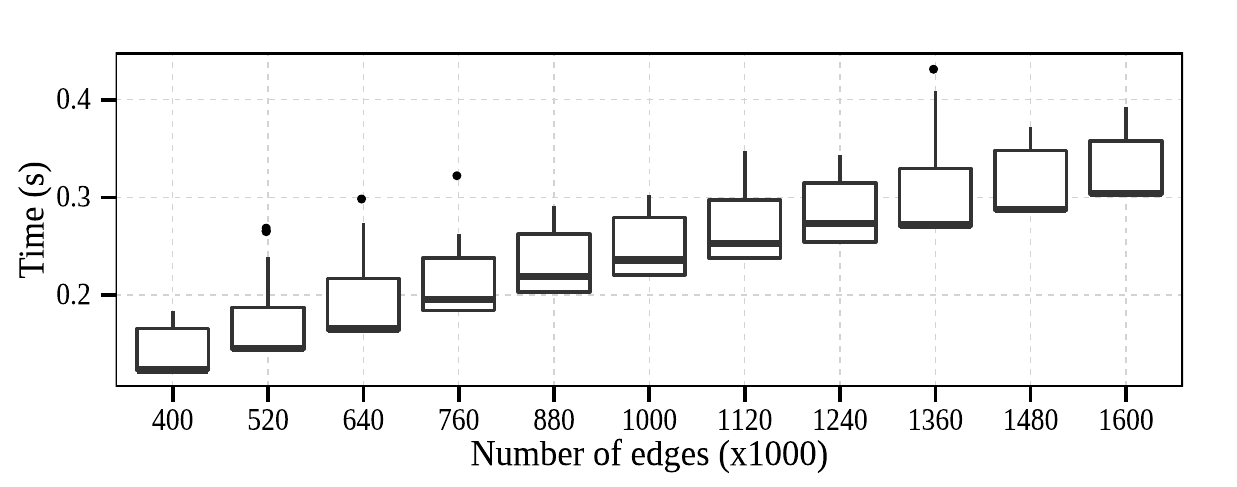}
    }
    \caption{
        Running time in seconds of iterations of the system in random
        networks.  (a) The input networks have 400\,000 edges and many different
        numbers of vertices. (b) 2\,000 vertices and many different numbers of edges.
    }
    \label{fig:time}
\end{figure}

    \cref{fig:time} depicts
    the running time of a single iteration of the system varying the number of
    vertices and edges, respectively.  With 10 independent runs, we measure the
    time for 30 iterations, totalizing 300 samples for each network size.
    We set $\lambda = 1$, two classes,
    and 5\% of labeled vertices.  Experiments were run on an
    Intel\textregistered~Core\texttrademark~i7 CPU 860 @ 2.80GHz with 16 GB RAM
    memory DDR3 @ 1333 MHz.  This experiment shows that the system
    runs in linear time as a function of the number of vertices and edges,
    which conforms our theoretical analysis.


\newcommand{\ThiagoModel}{Original Particle Competition\xspace}

\section{Computer Simulations}
\label{sec:simulation}


To study the stochastic system $\tilde{X}$ and the deterministic version $X$, we present experimental analyses
that concern their equivalence.  Additionally, we study the meaning of the parameters of
our learning model.  After that, we evaluate the model performance
using both artificial and synthetic datasets.  Then, we show the unfolding
process and the learning model on synthetic data.  Finally, we present the
simulation results for a well-known benchmark dataset and for a real
application on human activity and handwritten digits recognition.


\subsection{Experimental Analysis}
\label{sub:eanalysis}

In this section, we present an experiment that assesses the equivalence between
the unfolding results of both systems with an increasing initial number of
particles in \system{\tilde{X}}.

\newcommand{\OverlappingParam}{p}
\newcommand{\Labels}{\vec{y}}
\newcommand{\NLabels}{\card{\Labels}}
\newcommand{\Label}[1][i]{y_{#1}}
\newcommand{\Pref}[1][i]{d_{#1}}

The networks used for the analysis are generated by the following model:  a
complex network $G(\Labels, m, p)$ is constructed given a labeled vector
$\Labels$, a number $m>0$ of edges by vertex, and a weight $p \in [0,1]$ that
controls the preferential attachment between vertices of different classes.  The
resulting network contains $\NLabels$ vertices. For each \vertex{i}, $m$ edges
are randomly connected, with replacement. If $\Label[i] = \Label[j]$, the
preferential attachment weight is $1-p$; otherwise, the weight is $p$.  The
parameter $p$ is proportional to the overlap between classes.

\begin{figure}
    \centering
    \includegraphics[width=0.9\columnwidth]{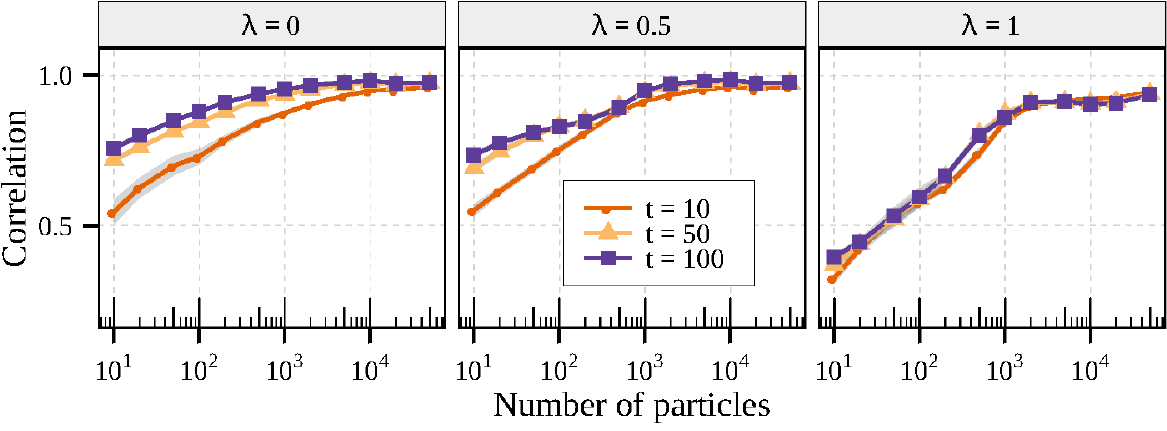}
    \caption{
        Proportionality simulation. Lines are the correlation measure between
        the \DominationName matrices of \systems{X} {\tilde{X}}, varying the
        initial number of active particles.  Values close to 1 indicate that the
        \DominationName matrices of both systems tend to be proportional.
    }
    \label{fig:corr}
\end{figure}

If there exists a positive constant $\kappa$ such that \[ \oDomination{c}(t) =
\kappa \Domination{c}(t)\mbox{,} \] both systems generate the same unfoldings.
To assess this proportionality, both systems are simulated in 10
different networks $G(\vec{y}, 3, 0.05)$, with $\card{\vec{y}} = 200$ vertices
arranged in two classes. The system's parameter is discretized in
$\SurvivalParam = \{0, 0.5, 1\}$.  Varying the total
number of initial particles, we set $\onpart[i](0) = \npart[i](0) \sim
\Degree{v_i}$ for all $c \in \{0, \dots, C\}$ and $i \in \{1, \dots,
\card{\VertexSet}\}$.

We consider the correlation between the
\DominationName matrices of \systems{X}{\tilde{X}}.  If the two matrices are
proportional, then they must be correlated.  Values of correlation close to 1
indicate the \DominationName matrices are proportional.
In \cref{fig:corr}, the correlation is depicted. As the number of initial
particles increases, the correlation approaches 1.
This result suggests that both systems generate the same unfolding
when the number of initial particles grows to infinity.


\subsection{Parameter Analysis}

The LCU model has two parameters apart from the network construction.  In this
section, we discuss their meaning.  To do so, the
learning model is applied in synthetic datasets whose data items are sampled
from a three dimensional knot torus $\vec{v}(\theta)$ with parametric curve
\[
    \begin{split}
        x(\theta) &= r(\theta) \cos{3 \theta},\\
        y(\theta) &= r(\theta) \sin{3 \theta},\\
        z(\theta) &= -\sin{4 \theta},
    \end{split}
\]
where $\theta \in [0, 2\pi]$ and $r(\theta) = 2 + \cos{4 \theta}$.

\begin{figure}
    \centering
    \includegraphics[width=0.8\columnwidth]{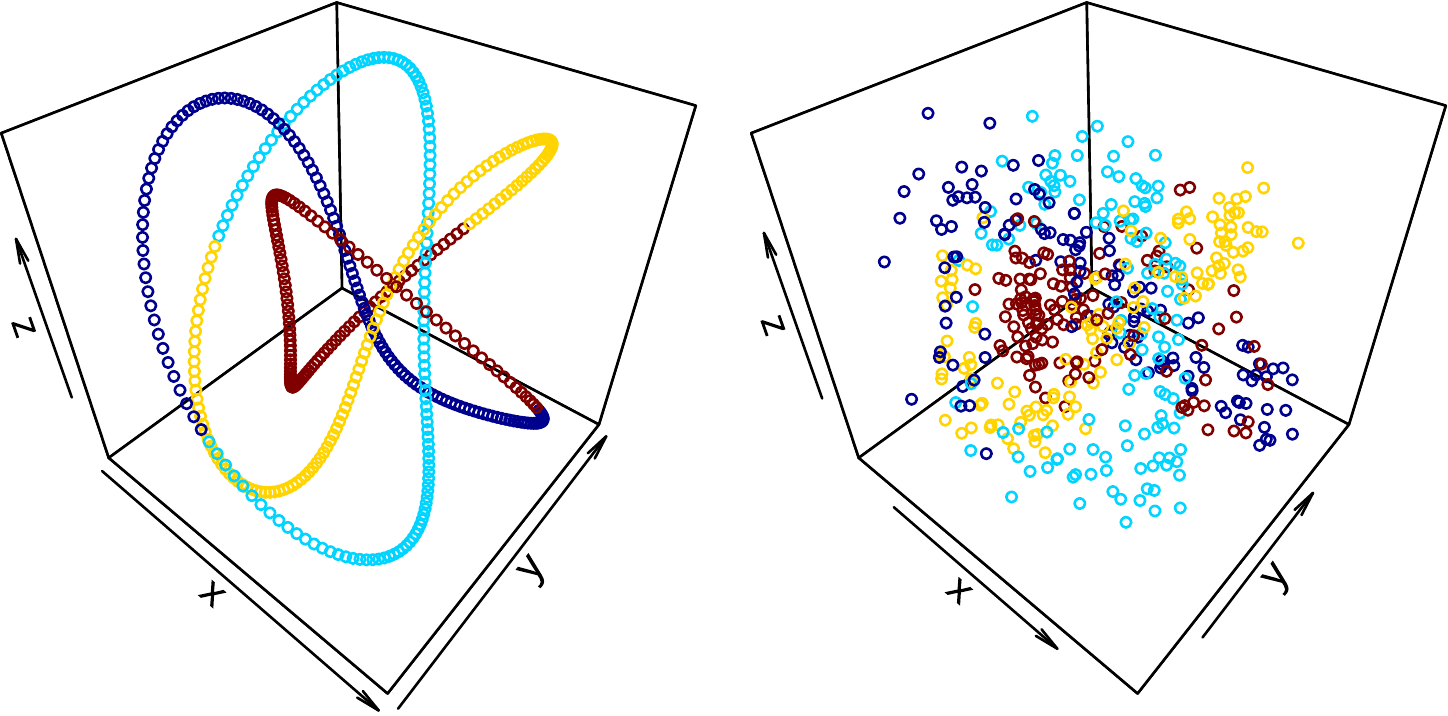}
    \caption{
            Three dimensional knot torus dataset with 500 samples without noise
            (left-hand side)
            and with noise (right-hand side).  Colors are the classes.
    }
    \label{fig:torus-data}
\end{figure}

We sampled 500 data items uniformly along the possible values of $\theta$.  We
randomly split the data items from 2 to 10 classes so that the samples with
adjacent $\theta$ belongs to the same class.  We also added to each sample a
random noise in each dimension with distribution $\mathcal{N}(0, \sigma)$ with
$\sigma = 0.25$ and $0.35$.  \cref{fig:torus-data} depicts an example of the
dataset with 4 classes with and without noise. Since the dataset has a complex
form, a small change of parameter value may generate different results.
Therefore, it is suitable to study the sensitivity of parameters.

We run the LCU model with parameters $\SurvivalParam \in \{0.25,
0.5, 0.75, 1\}$ and $\tau = 500$.  Finally, 30 unbiased sets of 40
labeled points are employed.  The $k$-NN is used for the network
construction with $k = \{4, 5, \dots, 10\}$.

Below, we discuss each parameter of the model.

\subsubsection{Discussion about the network construction parameter}

In our model, the input network must be simple (between any pair of vertices
there must exist at most one edge), unweighted, undirected, and
connected.  Besides these requirements, two vertices must be connected if their
data items are considered similar enough to the particular problem.
In our experiments, we use \kNN graph with Euclidean distance since it is
proved to approximate the low-dimensional manifold of point
set~\cite{Tenenbaum2000}.  The smaller the value of $k$, the better are
the results.

\subsubsection{Discussion about the system parameter}

\begin{figure}
    \centering
    \includegraphics[width=0.9\columnwidth]{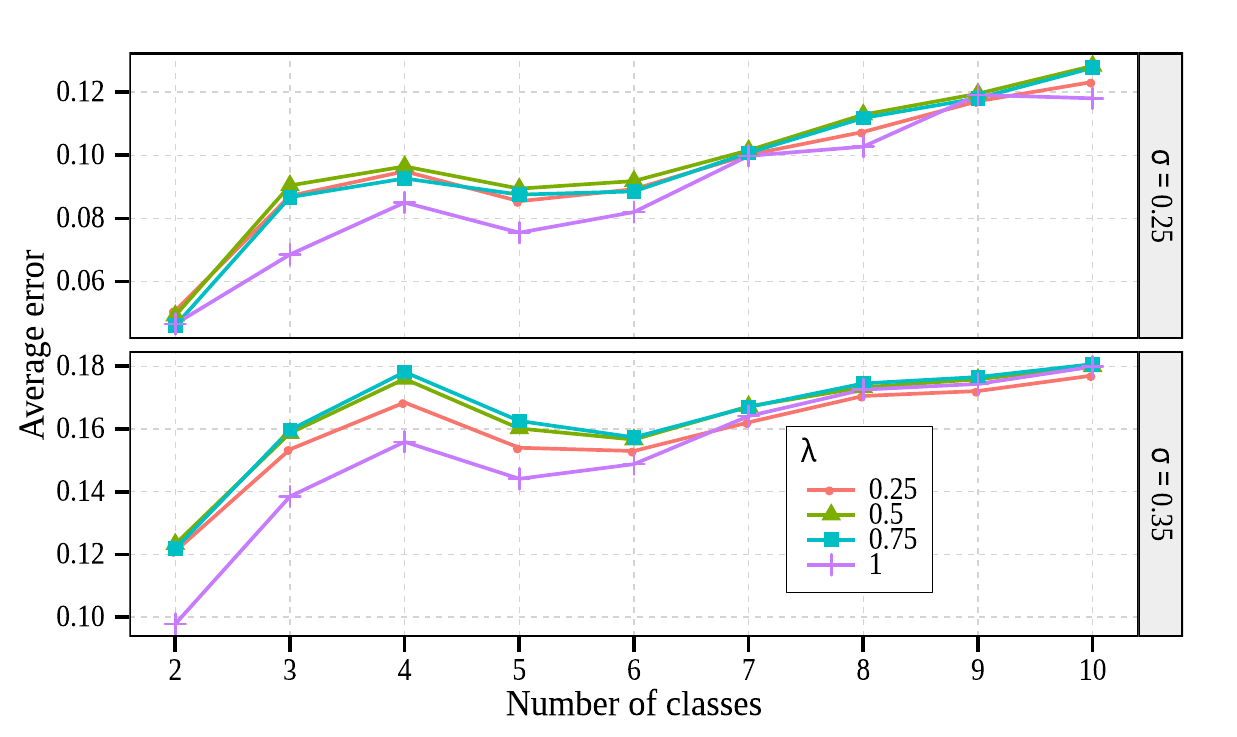}
    \caption{
            Average error of the proposed model for different numbers of classes
            in the problem.  Colors and shapes indicate the
            values of parameter $\lambda$.
    }
    \label{fig:torus-lambda}
\end{figure}

The LCU system has only one parameter: the competition parameter
$\SurvivalParam$.  This parameter defines the intensity of competition between
particles.  When $\SurvivalParam = 0$, particles
randomly walk the network, without competition.  As $\SurvivalParam$
approaches to 1, particles are more likely to compete and, consequently, to
be absorbed.  \cref{fig:torus-lambda}
depicts the average error of our method with different values of
$\SurvivalParam$.  Based on the figure, our model is not sensitive
to $\SurvivalParam$.  In general, we suggest setting
$\SurvivalParam = 1$ because of
better and more consistent classification than other values.

\subsubsection{Discussion about the system iteration stopping parameter}

The time
limit parameter $\tau$ controls when the simulation should stop;  it
must be at least as large as the diameter of the
network.  That way, it is guaranteed every edge to be visited by a
particle. Since the network diameter is usually a small value,
the simulation stops in few iterations.


\subsection{Simulations on Artificial Datasets}


\begin{figure}
    \centering
    \captionsetup[subfigure]{captionskip=-3pt}
    \subfloat[]{
        \includegraphics[width=0.9\columnwidth]{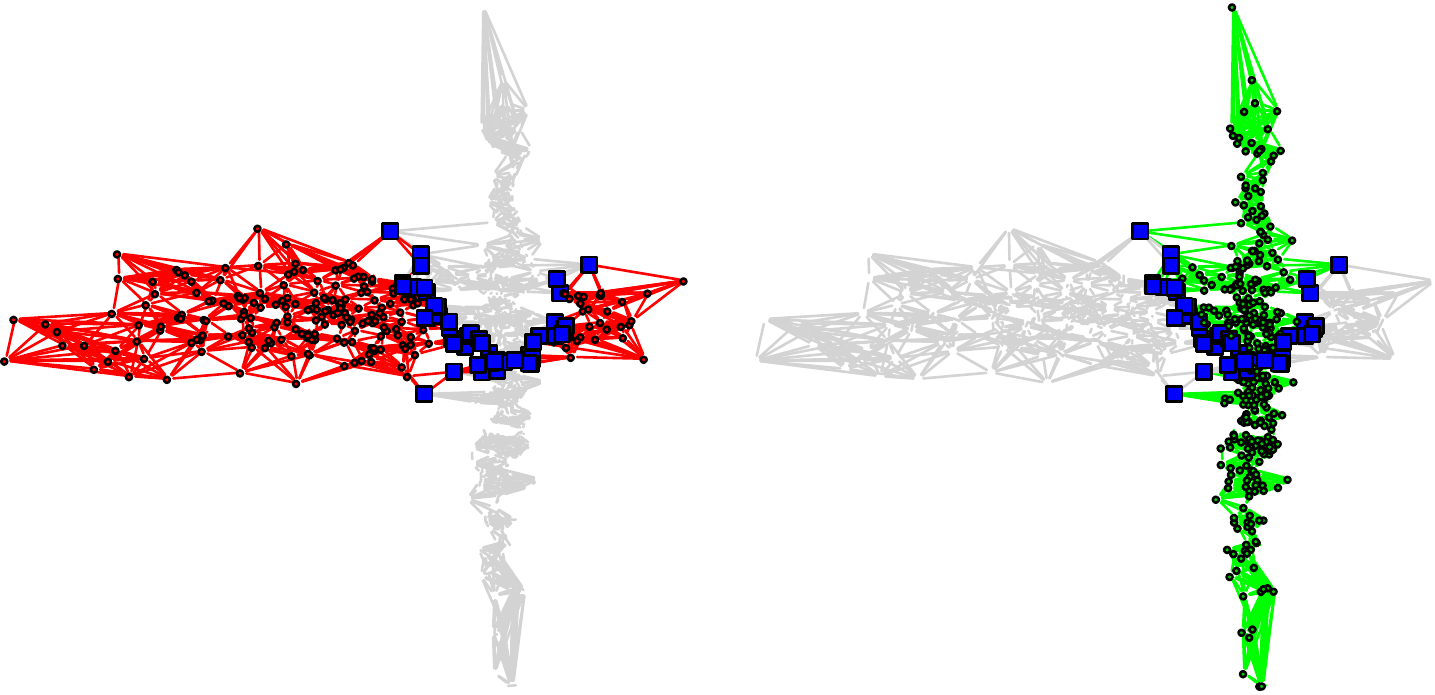}
        \label{fig:highleyman-unfolding-layout}
    }
    \hfill
    \subfloat[]{
        \includegraphics[width=0.9\columnwidth]{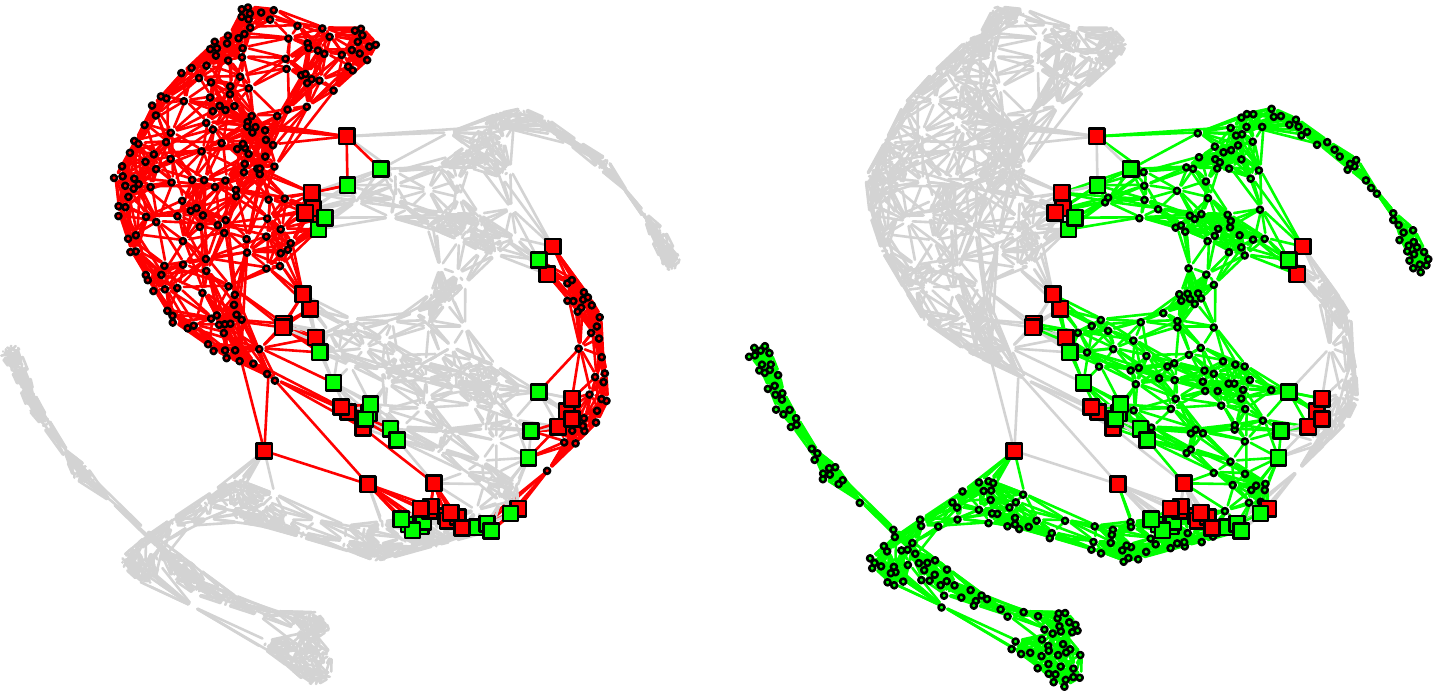}
        \label{fig:highleyman-unfolding}
    }
    \caption{
        Unfoldings generated by the proposed system at \time{t = 100} on
        Highleyman dataset.  Edges are colored according to the dominating class
        at the time.  Light gray edges stand for edges presented in the original
        network but not in the unfolding.  (a) Vertex position is imposed by
        the original data points and blue squares represent vertices
        connected in both unfoldings.  (b) Vertex position is not imposed by the
        original data points and color of the vertices are the result of
        the classification.
    }
\end{figure}

For better understanding the details of the \ShortModelName, in this subsection we
illustrate it using two synthetic datasets.  Each dataset has a different class
distribution---banana shape and Highleyman. (The datasets
are generated using the PRTools framework~\cite{Duin2007}.) The banana shape dataset
is uniformly distributed along specific shape, and then
superimposed on a normal distribution with the standard deviation of 1 along the
axes.  In Highleyman distribution, two classes are defined by bivariate normal
distributions with different parameters.  Because the datasets are not in a
network representation, we use \emph{k}-NN graph
construction method to transform them into respective network form. In the constructed network, a vertex
represents a datum, and it connects to its \emph{k} nearest
neighbors, determined by the Euclidean distance.  We set
$\SurvivalParam = 1$ for the simulation.

Firstly, the technique is tested on the Highleyman dataset.  Each class has 300
samples, of which 6 are labeled.
(We set $k = 10$ for the \emph{k}-NN algorithm.)  We
can observe that the labeled data points of the green class form a barrier to
samples of the red class.  The unfoldings $G^{\text{red}}(100)$ and
$G^{\text{green}}(100)$ are presented in \cref{fig:highleyman-unfolding-layout}.
In this figure, blue squares represent vertices that are connected by edges of
both unfoldings.  Besides of the labeled data of green class forming a barrier, the
constructed subnetworks are still connected---there is a single component
connecting all the vertices of the subnetwork.  It is better visualized in
\cref{fig:highleyman-unfolding}.  In this figure, the same unfoldings are
presented, but the positions of the vertices are not imposed by the original data.
Furthermore, colors of vertices in the figure indicate the
result of classification.
The overlapping data can be identified by the vertices that belong to two or
more unfoldings.
This result reveals that the competition system in edges provide more
information than the competition in vertices since it can identify the
overlapping vertices as part of the system, that is, without special treatments
or modifications.

\begin{figure}[!t]
    \centering
    \captionsetup[subfigure]{captionskip=-3pt}
    \subfloat[]{
        \label{fig:banana-network}
        \includegraphics[width=0.4\columnwidth]{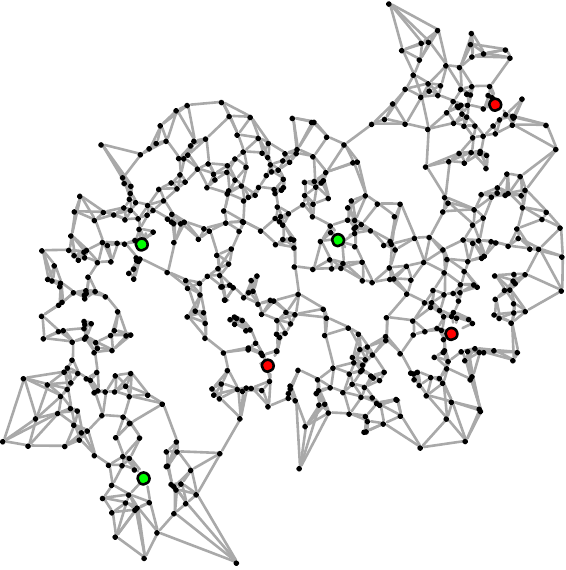}
    }
    \hfil
    \subfloat[]{
        \label{fig:banana-4}
        \includegraphics[width=0.4\columnwidth]{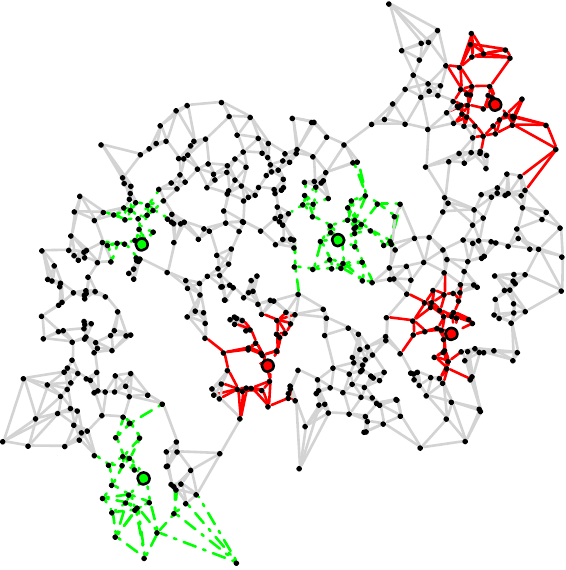}
    }
    \vspace{-0.1in}
    \hfil
    \subfloat[]{
        \label{fig:banana-20}
        \includegraphics[width=0.4\columnwidth]{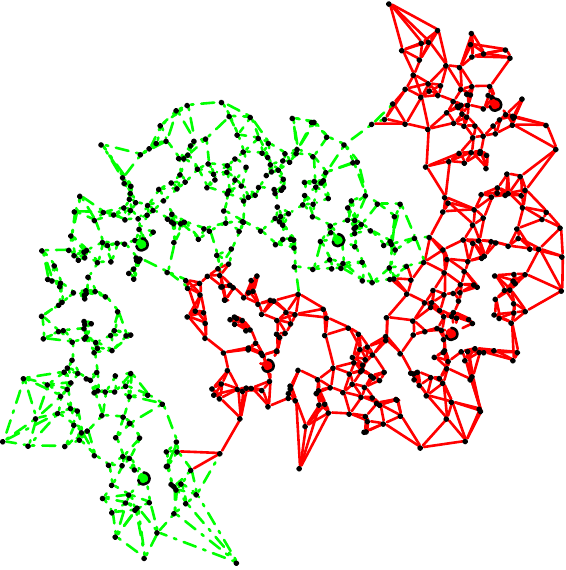}
    }
    \hfil
    \subfloat[]{
        \label{fig:banana-result}
        \includegraphics[width=0.4\columnwidth]{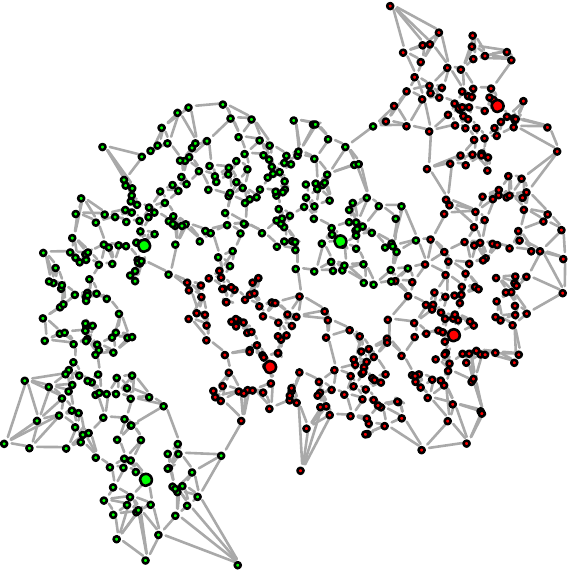}
    }
    \caption{
        System evolution on a banana-shaped distribution dataset. Red and green
        colors represent the two classes. Unlabeled points are black ones;
        labeled vertices are represented by larger and colored points. Edges are
        colored according to the dominating class at current iteration, where a light
        gray point stands for a vertex, which is not dominated yet.  (a) The network
        representation of the
        dataset at the beginning of the system. (b) and (c) System iteration
        at time 4
        and 20, respectively. (d) The result of the dataset classification.
    }
    \label{fig:banana-shaped}
\end{figure}

The last synthetic dataset has 600 samples equally split into two classes.  In
\cref{fig:banana-network}, the initial state of the system is illustrated, where
the dataset is represented by a network. (The network representation is obtained
by setting $k = 4$ for the \emph{k}-NN--graph construction.) At this stage, the
edges are not dominated by any of the classes.  Starting from this state,
labeled vertices (sources) generate particles that carry the label of the
sources.  Though the particles are not shown,
\cref{fig:banana-4,fig:banana-20} are snapshots of
the system evolution---at time 4 and 20---where each edge is colored by its
dominating class at that iteration. In these illustrations, a solid red
line stands for an \edge{i}{j} that $\Domination{\text{red}} +
\Domination[ji]{\text{red}} > \Domination{\text{green}} +
\Domination[ji]{\text{green}}$, while a dashed green line stands for the
opposite. When $\Domination{\text{red}} + \Domination[ji]{\text{red}} =
\Domination{\text{green}} + \Domination[ji]{\text{green}}$, an edge is drawn in
a solid light gray line. As expected, edges close to sources are dominated
initially, and farther edges are progressively dominated. At time 20,
\cref{fig:banana-20}, every edge has been dominated, and the edge domination does
not change anymore. \cref{fig:banana-result} shows dataset classification
following the system result.  In this example, with 1\% of points in the labeled
set, the technique can correctly identify the pattern formed by each
class.  Both results are satisfactory, reinforcing the ability of
the technique of learning arbitrary class distributions.


\subsection{Simulations on Benchmark Datasets}

\newcommand{\BParam}[4]{#1&#2} 

\begin{table}[!t]
\caption{Test Errors (\%) with Standard Deviation and the Best Parameters}
\label{tbl:sd-and-param}
\begin{adjustbox}{width={\columnwidth}}
\begin{tabular}{lrrlrrl}
 \toprule
 & \bf 10 labeled & $k$ & $\SurvivalParam$ %
 & \bf 100 labeled & $k$ & $\SurvivalParam$%
\\ \midrule
\bf g241c   & \rvaluesd{42.90}{4.33} & \BParam{10}{0.25}{0}{1}
            & \rvaluesd{30.03}{2.18} & \BParam{10}{0.875}{0}{1} \\
\bf g241n   & \rvaluesd{46.94}{3.93} & \BParam{9}{0}{0}{1}
            & \rvaluesd{36.08}{6.32} & \BParam{9}{0}{0}{1} \\ 
\bf Digit1  & \rvaluesd{ 4.93}{2.63} & \BParam{5}{0.75}{0}{3}
            & \rvaluesd{ 1.51}{0.31} & \BParam{6}{0.625}{0}{1} \\ 
\bf USPS    & \rvaluesd{15.65}{3.81} & \BParam{3}{1}{0.25}{3}
            & \rvaluesd{ 8.36}{2.92} & \BParam{3}{1}{0.25}{2} \\ 
\bf COIL    & \rvaluesd{59.96}{6.13} & \BParam{3}{0.625}{0}{1}
            & \rvaluesd{13.73}{2.91} & \BParam{3}{0}{0.25}{1} \\ 
\bf BCI     & \rvaluesd{47.56}{1.80} & \BParam{9}{1}{0}{2}
            & \rvaluesd{34.68}{2.26} & \BParam{3}{0.25}{0}{1} \\ 

\bf Text    & \rvaluesd{29.71}{3.53} & \BParam{9}{0.875}{0}{3}
            & \rvaluesd{22.41}{1.74} & \BParam{10}{0.75}{0}{1}  
\\ \bottomrule
\end{tabular}
\end{adjustbox}
\end{table}


%

\begin{table}[!t]
\renewcommand{\arraystretch}{1.0}
\caption{Test Errors (\%) with 10 Labeled Training Points}
\label{tbl:10}
\begin{adjustbox}{width={\columnwidth}}
\centering
\begin{tabular}{lrrrrrrrr}
 \toprule
& \bf g241c & \bf g241d & \bf Digit1 & \bf USPS & \bf COIL
& \bf BCI & \bf Text & \bf Avg. Rank 
\\ \midrule
{\bf 1-NN}
    & \rvalue{47.88}{} 
    & \rvalue{46.72}{} 
    & \rvalue{13.65}{} 
    & \rvalue{16.66}{} 
    & \rvalue{63.36}{} 
    & \rvalue{49.00}{} 
    & \rvalue{38.12}{} 
    & \rvalue{ 9.3 }{} 
\\
{\bf SVM}
    & \rvalue{47.32}{} 
    & \rvalue{46.66}{} 
    & \rvalue{30.60}{} 
    & \rvalue{20.03}{} 
    & \rvalue{68.86}{} 
    & \rvalue{49.85}{} 
    & \rvalue{45.37}{} 
    & \rvalue{13.0 }{} 
\\
{\bf MVU + 1-NN}
    & \rvalue{47.15}{} 
    & \rvalue{45.56}{} 
    & \rvalue{14.42}{} 
    & \rvalue{23.34}{} 
    & \rvalue{62.62}{} 
    & \rvalue{47.95}{} 
    & \rvalue{45.32}{} 
    & \rvalue{ 9.3 }{} 
\\
{\bf LEM + 1-NN}
    & \rvalue{44.05}{} 
    & \rvalue{43.22}{} 
    & \rvalue{23.47}{} 
    & \rvalue{19.82}{} 
    & \rvalue{65.91}{} 
    & \rvalue{48.74}{} 
    & \rvalue{39.44}{} 
    & \rvalue{ 9.1 }{} 
\\
{\bf QC + CMR}
    & \rvalue{39.96}{} 
    & \rvalue{46.55}{} 
    & \rvalue{ 9.80}{} 
    & \rvalue{13.61}{} 
    & \rvalue{59.63}{} 
    & \rvalue{50.36}{} 
    & \rvalue{40.79}{} 
    & \rvalue{ 6.9 }{} 
\\
{\bf Discrete Reg.}
    & \rvalue{49.59}{} 
    & \rvalue{49.05}{} 
    & \rvalue{12.64}{} 
    & \rvalue{16.07}{} 
    & \rvalue{63.38}{} 
    & \rvalue{49.51}{} 
    & \rvalue{40.37}{} 
    & \rvalue{10.4 }{} 
\\
{\bf TSVM}
    & \rvalue{24.71}{} 
    & \rvalue{50.08}{} 
    & \rvalue{17.77}{} 
    & \rvalue{25.20}{} 
    & \rvalue{67.50}{} 
    & \rvalue{49.15}{} 
    & \rvalue{31.21}{} 
    & \rvalue{10.0 }{} 
\\
{\bf Cluster--Kernel}
    & \rvalue{48.28}{} 
    & \rvalue{42.05}{} 
    & \rvalue{18.73}{} 
    & \rvalue{19.41}{} 
    & \rvalue{67.32}{} 
    & \rvalue{48.31}{} 
    & \rvalue{42.72}{} 
    & \rvalue{10.1 }{} 
\\
{\bf LDS}
    & \rvalue{28.85}{} 
    & \rvalue{50.63}{} 
    & \rvalue{15.63}{} 
    & \rvalue{17.57}{} 
    & \rvalue{61.90}{} 
    & \rvalue{49.27}{} 
    & \rvalue{27.15}{} 
    & \rvalue{ 8.0 }{} 
\\
{\bf Laplacian RLS}
    & \rvalue{43.85}{} 
    & \rvalue{45.68}{} 
    & \rvalue{ 5.44}{} 
    & \rvalue{18.99}{} 
    & \rvalue{54.54}{} 
    & \rvalue{48.97}{} 
    & \rvalue{33.68}{} 
    & \rvalue{ 5.9 }{} 
\\
{\bf LGC}
    & \rvalue{45.82}{} 
    & \rvalue{44.09}{} 
    & \rvalue{ 9.89}{} 
    & \rvalue{ 9.03}{} 
    & \rvalue{63.45}{} 
    & \rvalue{47.09}{} 
    & \rvalue{46.83}{} 
    & \rvalue{ 6.9 }{} 
\\
{\bf LP}
    & \rvalue{42.61}{} 
    & \rvalue{41.93}{} 
    & \rvalue{11.31}{} 
    & \rvalue{14.83}{} 
    & \rvalue{55.82}{} 
    & \rvalue{46.37}{} 
    & \rvalue{49.53}{} 
    & \rvalue{ 5.1 }{} 
\\
{\bf LNP}
    & \rvalue{47.82}{} 
    & \rvalue{46.24}{} 
    & \rvalue{ 8.58}{} 
    & \rvalue{17.87}{} 
    & \rvalue{55.50}{} 
    & \rvalue{47.65}{} 
    & \rvalue{41.06}{} 
    & \rvalue{ 7.1 }{} 
\\
{\bf \ThiagoModel}
    & \rvalue{41.17}{} 
    & \rvalue{43.51}{} 
    & \rvalue{ 8.10}{} 
    & \rvalue{15.69}{} 
    & \rvalue{54.18}{} 
    & \rvalue{48.00}{} 
    & \rvalue{34.84}{} 
    & \rvalue{ 4.0 } {} 
\\ 
{\bf \itshape \SimulationModelName}
    & \rvalue{42.90}{} 
    & \rvalue{46.94}{} 
    & \rvalue{ 4.93}{} 
    & \rvalue{15.65}{} 
    & \rvalue{59.96}{} 
    & \rvalue{47.56}{} 
    & \rvalue{29.71}{} 
    & \rvalue{ 4.9 }{} 
\\
\bottomrule
\end{tabular}
\end{adjustbox}
\end{table}


\begin{table}[!t]
\renewcommand{\arraystretch}{1.0}
\caption{Test Errors (\%) with 100 Labeled Training Points}
\label{tbl:100}
\begin{adjustbox}{width={\columnwidth}}
\centering
\begin{tabular}{lrrrrrrrr}
\toprule
& \bf g241c & \bf g241d & \bf Digit1 & \bf USPS & \bf COIL
& \bf BCI & \bf Text & \bf Avg. Rank
\\ \midrule
\bf 1-NN
    & \rvalue{43.93}{} 
    & \rvalue{42.45}{} 
    & \rvalue{ 3.89}{} 
    & \rvalue{ 5.81}{} 
    & \rvalue{17.35}{} 
    & \rvalue{48.67}{} 
    & \rvalue{30.11}{} 
    & \rvalue{11.4 }{} 
\\
\bf SVM
    & \rvalue{23.11}{} 
    & \rvalue{24.64}{} 
    & \rvalue{ 5.53}{} 
    & \rvalue{ 9.75}{} 
    & \rvalue{22.93}{} 
    & \rvalue{34.31}{} 
    & \rvalue{26.45}{} 
    & \rvalue{ 8.1 }{} 
\\
\bf MVU + 1-NN
    & \rvalue{43.01}{} 
    & \rvalue{38.20}{} 
    & \rvalue{ 2.83}{} 
    & \rvalue{ 6.50}{} 
    & \rvalue{28.71}{} 
    & \rvalue{47.89}{} 
    & \rvalue{32.83}{} 
    & \rvalue{10.6 }{} 
\\
\bf LEM + 1-NN
    & \rvalue{40.28}{} 
    & \rvalue{37.49}{} 
    & \rvalue{ 6.12}{} 
    & \rvalue{ 7.64}{} 
    & \rvalue{23.27}{} 
    & \rvalue{44.83}{} 
    & \rvalue{30.77}{} 
    & \rvalue{10.9 }{} 
\\
\bf QC + CMR
    & \rvalue{22.05}{} 
    & \rvalue{28.20}{} 
    & \rvalue{ 3.15}{} 
    & \rvalue{ 6.36}{} 
    & \rvalue{10.03}{} 
    & \rvalue{46.22}{} 
    & \rvalue{25.71}{} 
    & \rvalue{ 6.6 }{} 
\\
\bf Discrete Reg.
    & \rvalue{43.65}{} 
    & \rvalue{41.65}{} 
    & \rvalue{ 2.77}{} 
    & \rvalue{ 4.68}{} 
    & \rvalue{ 9.61}{} 
    & \rvalue{47.67}{} 
    & \rvalue{24.00}{} 
    & \rvalue{ 7.1 }{} 
\\
\bf TSVM
    & \rvalue{18.46}{} 
    & \rvalue{22.42}{} 
    & \rvalue{ 6.15}{} 
    & \rvalue{ 9.77}{} 
    & \rvalue{25.80}{} 
    & \rvalue{33.25}{} 
    & \rvalue{24.52}{} 
    & \rvalue{ 7.7 }{} 
\\
\bf Cluster-Kernel
    & \rvalue{13.49}{} 
    & \rvalue{ 4.95}{} 
    & \rvalue{ 3.79}{} 
    & \rvalue{ 9.68}{} 
    & \rvalue{21.99}{} 
    & \rvalue{35.17}{} 
    & \rvalue{34.28}{} 
    & \rvalue{ 7.4 }{} 
\\
\bf LDS
    & \rvalue{18.04}{} 
    & \rvalue{23.74}{} 
    & \rvalue{ 3.46}{} 
    & \rvalue{ 4.96}{} 
    & \rvalue{13.72}{} 
    & \rvalue{43.97}{} 
    & \rvalue{23.15}{} 
    & \rvalue{ 5.4 }{} 
\\
\bf Laplacian RLS
    & \rvalue{24.36}{} 
    & \rvalue{26.46}{} 
    & \rvalue{ 2.92}{} 
    & \rvalue{ 4.68}{} 
    & \rvalue{11.92}{} 
    & \rvalue{31.36}{} 
    & \rvalue{23.57}{} 
    & \rvalue{ 4.4 }{} 
\\
\bf LGC
    & \rvalue{41.64}{} 
    & \rvalue{40.08}{} 
    & \rvalue{ 2.72}{} 
    & \rvalue{ 3.68}{} 
    & \rvalue{45.55}{} 
    & \rvalue{43.50}{} 
    & \rvalue{56.83}{} 
    & \rvalue{ 9.3 }{} 
\\
\bf LP
    & \rvalue{30.39}{} 
    & \rvalue{29.22}{} 
    & \rvalue{ 3.05}{} 
    & \rvalue{ 6.98}{} 
    & \rvalue{11.14}{} 
    & \rvalue{42.69}{} 
    & \rvalue{40.79}{} 
    & \rvalue{ 8.3 }{} 
\\
\bf LNP
    & \rvalue{44.13}{} 
    & \rvalue{38.30}{} 
    & \rvalue{ 3.27}{} 
    & \rvalue{17.22}{} 
    & \rvalue{11.01}{} 
    & \rvalue{46.22}{} 
    & \rvalue{38.45}{} 
    & \rvalue{11.4 }{} 
\\
\bf \ThiagoModel
    & \rvalue{21.41}{} 
    & \rvalue{25.85}{} 
    & \rvalue{ 3.11}{} 
    & \rvalue{ 4.82}{} 
    & \rvalue{10.94}{} 
    & \rvalue{41.57}{} 
    & \rvalue{27.92}{} 
    & \rvalue{ 5.3 }{} 
\\ 
\bf \itshape \SimulationModelName
    & \rvalue{30.03}{} 
    & \rvalue{36.08}{} 
    & \rvalue{ 1.51}{} 
    & \rvalue{ 8.36}{} 
    & \rvalue{13.73}{} 
    & \rvalue{34.68}{} 
    & \rvalue{22.41}{} 
    & \rvalue{ 6.0 }{} 
\\
\bottomrule
\end{tabular}
\end{adjustbox}
\end{table}

We compare our model with 14 semi-supervised techniques tested on Chapelle's
benchmark~\cite{Chapelle2006}.  The benchmark is formed by seven datasets that
have 1500 data points, except for BCI that has 400 points. The datasets are
described in \cite{Chapelle2006}.

For each dataset, 24 distinct, unbiased sets (splits) of labeled points are
provided within the benchmark.  Half of the splits are formed by 10 labeled
points and the other half by 100 labeled points. The author of the benchmark
ensured that each split contains at least one data point of each class. The
result is the average test error---the proportion of data points incorrectly
labeled---over the splits.  We compare our results to the ones obtained by the following
techniques: 1-Nearest Neighbors (1-NN), Support Vector Machines (SVM), Maximum
variance unfolding (MVU + 1-NN), Laplacian eigenmaps (LEM + 1-NN), Quadratic
criterion and class mass regularization (QC + CMR), Discrete regularization
(Discrete reg.), Transductive support vector machines (TSVM), Cluster kernels
(Cluster-Kernel), Low-density separation (LDS), Laplacian regularized least
squares (Laplacian RLS), Local and global consistency (LGC), Label propagation
(LP), Linear neighborhood propagation (LNP), and Network-Based Stochastic
Semisupervised Learning (Vertex Domination), The simulation results are
collected from~\cite{Chapelle2006}, except for LGC, LP, LNP, and \ThiagoModel
that are found in~\cite{Silva2012}.



For the simulation of the \ShortModelName, we discretize the interval of the parameter in
$\SurvivalParam = \left\{0, 0.125, \dots, 1\right\}$. Also, we vary the
\emph{k}-NN parameter $k \in \left\{1, 2, \dots, 10\right\}$. We tested every
combination of $k$ and $\SurvivalParam$.  Moreover, we
fix $\tau = 1000$.
In \cref{tbl:sd-and-param}, we present the results with the standard
deviation over the splits along with the best combination of parameters
that generated the best accuracy result.

The test error comparison for 10 labeled points are shown in \cref{tbl:10};
comparison for 100 labeled points are in \cref{tbl:100}. Apart from each dataset,
the last column is the average performance rank of a technique over the
datasets.
A ranking arranges the methods under comparison by test error rate in ascending
order. For a single dataset, we assign rank 1 for the method with the lowest
average test error on that dataset, then rank 2 for the method with the
second lowest test error, and so on. The average ranking is the average value of
the rankings of the method on all the datasets. The smaller the ranking score,
the better the method has performed.

From the average rank column, the LCU
technique is not the best ranked, but it is in the best group of
techniques in both 10 labeled and 100 labeled cases.

We statistically compare the results presented in \cref{tbl:10,tbl:100}.
For all tests we set a significance level of $5\%$. First,
we use a test based on the average rank of each method to evaluate the
null hypothesis that all the techniques are equivalent.  With the Friedman
test~\cite{Hollander1999},  there is statistically significant difference
between the rank of the techniques


\newcommand{\pvalue}[1]{}

Since the Friedman test result reports statistical significance,
we use the Wilcoxon signed-rank test~\cite{Hollander1999}.
In this pairwise difference test, we
test for the null hypothesis that the first technique has greater or equal error
results than the second.
If rejected at a 5\% significance level, then we say the first technique is
superior to the second. %
By analyzing results for 10 and 100 labeled points together, we conclude that
our technique is superior to 1-NN, LEM + 1-NN, and MVU + 1-NN.
Examining separately, for 10 labeled points, our method is also superior to
discrete regularization, cluster kernel,
and SVM. For 100 labeled points, it is also superior to LNP and LGC;
whereas Laplacian RLS is superior to ours.
%



\subsection{Simulations on Human Activity Dataset}

\begin{table*}[ht!]
\caption{Performance Comparison in the Human Activity Recognition Using
Smartphones Dataset}
\label{tbl:hapt-vs}
\begin{adjustbox}{width={\textwidth}}
\begin{tabular}{l|ccccccccc|ccc}
 \toprule
& \multicolumn{9}{c|}{Labeled Component Unfolding} & \multicolumn{3}{c}{SVM}\\
& \multicolumn{3}{c}{5\% labeled} & \multicolumn{3}{c}{10\% labeled} & \multicolumn{3}{c|}{20\% labeled} & \multicolumn{3}{c}{$\approx$70\% labeled \cite{Anguita2013dataset}}\\
& Precision & Recall & F Score & Precision & Recall & F Score & Precision & Recall & F Score & Precision & Recall & F Score \\
 \midrule
\bf WK & .984 $\pm$ .013 & .941 $\pm$ .030 & .962 $\pm$ .016 & .992 $\pm$ .004 & .985 $\pm$ .011 & .989 $\pm$ .006 & .994 $\pm$ .002 & .997 $\pm$ .001 & \bf .995 $\pm$ .001 & .957 & .992 & .974\\
\bf WU & .981 $\pm$ .009 & .935 $\pm$ .026 & .957 $\pm$ .015 & .988 $\pm$ .008 & .961 $\pm$ .013 & .974 $\pm$ .008 & .991 $\pm$ .003 & .981 $\pm$ .006 & \bf .986 $\pm$ .004 & .980 & .958 & .969\\
\bf WD & .987 $\pm$ .017 & .901 $\pm$ .016 & .942 $\pm$ .011 & .994 $\pm$ .008 & .918 $\pm$ .011 & .955 $\pm$ .007 & .998 $\pm$ .001 & .945 $\pm$ .008 & .971 $\pm$ .004 & .988 & .976 & \bf .982\\
\bf ST & .864 $\pm$ .034 & .698 $\pm$ .049 & .770 $\pm$ .022 & .883 $\pm$ .015 & .743 $\pm$ .039 & .806 $\pm$ .020 & .905 $\pm$ .014 & .814 $\pm$ .015 & .857 $\pm$ .006 & .969 & .880 & \bf .922\\
\bf SD & .840 $\pm$ .024 & .842 $\pm$ .053 & .839 $\pm$ .017 & .870 $\pm$ .023 & .844 $\pm$ .022 & .856 $\pm$ .006 & .896 $\pm$ .013 & .872 $\pm$ .021 & .884 $\pm$ .009 & .901 & .974 & \bf .936\\
\bf LD & .996 $\pm$ .002 & .999 $\pm$ .000 & .998 $\pm$ .001 & .997 $\pm$ .001 & .999 $\pm$ .000 & .998 $\pm$ .000 & .997 $\pm$ .001 & .999 $\pm$ .000 & .998 $\pm$ .000 & 1.000 & 1.000 & \bf 1.000\\
 \bottomrule
\end{tabular}
\end{adjustbox}
\end{table*}

The Human Activity Recognition Using Smartphones~\cite{Anguita2013dataset}
dataset comprises of 10299 data samples.  Each sample matches 561 features
extracted from motion sensors attached to a person during a time window.  Each
person performed six activities which are target labels in the dataset---walking
(WK), walking upstairs (WU), walking downstairs (WD), sitting (ST), standing
(SD), and laying down (LD).

We use \kNN with $k=7$ for the dataset network representation once it is the
smallest value that generates a connected network.  The parameters are fixed in
$\lambda = 1$ and $\tau = 1000$.  We compare our results with the ones published
in \cite{Anguita2013dataset}, splitting the problem into six binary
classification tasks.

\cref{tbl:hapt-vs} summarises the results.  For our technique, we provide the
precision, recall and F Score using 5\%, 10\%, and 20\% of labeled samples.  We
average the results of 10 independent labeled set for each configuration. We
also provide the original results from \cite{Anguita2013dataset} using SVM with
approximately 70\% of labeled samples.  Our technique performs as well as SVM
\emph{using far fewer labeled samples} and using the suggested parameter set.  Such a
feature is quite attractive because it may represent a big saving in money or
efforts when involving manually data labeling in semi-supervised learning.


\subsection{Simulations on MNIST Dataset}

\begin{table}[t!]
\caption{Test Errors (\%) in the MNIST Dataset}
\label{tbl:mnist-vs}
\centering
\begin{threeparttable}
\begin{tabular}{c c c}
 \toprule
 Method & 100 labeled & 1000 labeled \\
 \midrule
 \emph{LCU} & $10.62 \pm 1.91$ & $6.31 \pm 0.46$ \\
 TSVM\textsuperscript{*} \cite{Collobert2006} & $16.81$ & $5.65$ \\
 Embed NN\textsuperscript{*} \cite{Weston2012} & $16.86$ & $8.52$ \\
 Embed CNN\textsuperscript{*} \cite{Weston2012} & $7.75$ & $3.82$ \\
 \bottomrule
\end{tabular}
\begin{tablenotes}
\item \textsuperscript{*} The comparison is biased since the results from
  \cite{Collobert2006,Weston2012} rely on a single and unique labeled set.  See
  text for more details.
\end{tablenotes}
\end{threeparttable}
\end{table}

The MNIST dataset comprises 70,000 examples of handwritten digits.  All digits
have been size-normalized and centered in a fixed-size image.  In a supervised
learning setting, this dataset is split into two sets: 60,000 examples for
training and 10,000 for testing.

To adapt the dataset to a semi-supervised learning problem, we use a setting
similar to \cite{Collobert2006,Weston2012}: the labeled input data items are selected from
the training set, and the unlabeled ones from the test set.  Although we do not
use a validation set, \cite{Collobert2006} and \cite{Weston2012} use an additional set of
at least 1,000 labeled samples for parameter tuning.

The network representation is obtained from the images without preprocessing. We
use the Euclidean distance between items and $k = 3$ to construct the \kNN
network.  Similarly to the previous experiment, the value of $k$ is the smallest
value that generates a connected network.  The parameters are fixed in $\lambda
= 0.9$ and $\tau = 500$.

\cref{tbl:mnist-vs} compares the error rate of our method to other
semi-supervised techniques.  To the best of our knowledge, we could not find many
papers with experiments under the same semi-supervised settings for the MNIST
dataset.  Due to such available results in the literature, we sought to carry
our experiments with the input as similar as possible to the compared results.
However, a single sampling with as less as 1,000 labeled samples out of a set of
60,000 images most probably results in a biased accuracy result, we opt to
average results from 15 labeled sets for each parameter setting.

Our model performs well even without preprocessing and validation set.  This
result indicates that besides the simplicity of the constructed network, our
learning system can obtain enough knowledge from data.


\section{Conclusion}
\label{sec:conclusion}

We have presented a transductive semi-supervised learning technique based on a
vertex-edge dynamical system on complex networks. First, the input data is mapped into a
network. Then, the proposed \ModelNameFirst runs on this network. At this
stage, particles compete for edges in the network. When a particle passes
through an edge, it increases its class dominance over the edge while decreasing
other classes' dominance.  Three dynamics---\WDynamicsName,
absorption and production---provide a biologically inspired scenario of
competition and cooperation.  Then, labels are assigned according to the
dominant class over the edges.  As a result,
the system unfolds the original network by grouping edges dominated by the same
class.  Finally, we employ the unfoldings to classify unlabeled data.
Furthermore, rigorous studies have been done on the novel \ShortModelName.

The deterministic system implementation brings advantages over its stochastic
counterpart.  The time complexity of the deterministic one does not depend on
the number of particles, so we are benefited from better results when considering
a continuously varying number of initial particles.  Besides, the \ShortModelName allows a stable transductive semi-supervised learning
technique with a subquadratic order of complexity. Computer simulations show the
proposed technique achieves a good classification accuracy and it is suitable for
situations where a small number of labeled samples are available.
Another interesting feature of the proposed model is that it directly provides
the overlapping information of each vertex or a subset of vertices.

As future works, we would like to investigate the mathematical property of the
\ShortModelName on directed or weighted networks. Besides of this, it is
interesting to improve the runtime further via network sampling methods or
estimation methods. In this way, the model will be suitable to be applied to
process large enough datasets or streaming data. Another interesting research
is to treat the labels on edges instead of nodes in a semi-supervised learning
environment.

\bibliographystyle{IEEEtran}
\bibliography{IEEEabrv,journal}

\end{document}